\documentclass{article}
\usepackage{preprint,times}

\usepackage{amsmath,amsfonts,bm}

\def\eqref#1{equation~\ref{#1}}
\def\1{\bm{1}}

\DeclareMathAlphabet{\mathsfit}{\encodingdefault}{\sfdefault}{m}{sl}
\SetMathAlphabet{\mathsfit}{bold}{\encodingdefault}{\sfdefault}{bx}{n}

\usepackage{hyperref}
\usepackage{url}
\usepackage{microtype}
\usepackage{graphicx}
\usepackage{enumitem}
\usepackage{dsfont}
\usepackage{booktabs}
\usepackage{amsmath}
\usepackage{caption}
\usepackage{amssymb}
\usepackage{pifont}
\usepackage{makecell}
\newcommand{\cmark}{\ding{51}}
\newcommand{\xmark}{\ding{55}}
\usepackage{algorithm}
\usepackage{algpseudocode}
\usepackage{newunicodechar}
\usepackage{wrapfig}
\usepackage{xcolor}
\usepackage{soul}

\definecolor{markgreen}{RGB}{34,139,34}
\definecolor{markred}{RGB}{200,30,30}
\providecommand{\gcmark}{\textcolor{markgreen}{\cmark}}
\providecommand{\rxmark}{\textcolor{markred}{\xmark}}

\usepackage[most]{tcolorbox}
\newtcolorbox{promptbox}[2][]{enhanced, breakable, sharp corners=downhill, arc=1.5pt,
  colback=gray!6, colframe=gray!45, boxrule=0.4pt,
  colbacktitle=gray!18, coltitle=black, fonttitle=\small\bfseries, title={#2},
  toptitle=2pt, bottomtitle=2pt, left=6pt, right=6pt, top=5pt, bottom=6pt,
  fontupper=\small, parbox=false, before skip=10pt, after skip=10pt, #1}

 \usepackage{booktabs}
 \usepackage{arydshln}
 \usepackage{algorithm,algpseudocode,multicol}
 \usepackage{tabularx}
\newcommand{\AuthorOne}{Guanghan Ning}
\newcommand{\AuthorTwo}{Ping Liu}
\newcommand{\AuthorThree}{Linyi Li}
\newcommand{\AuthorFour}{Huangjie Zheng}
\newcommand{\AuthorFive}{Arjun Neervannan}
\newcommand{\AuthorSix}{Huu Nguyen}
\newcommand{\AuthorSeven}{Michael Sklar}
\newcommand{\AuthorEight}{Deniz Zorlu}
\newcommand{\AuthorNine}{Nicolai Ouporov}
\newcommand{\AffilOne}{Fleet AI}
\newcommand{\AffilTwo}{University of Nevada, Reno}
\newcommand{\AffilThree}{Simon Fraser University}
\newcommand{\AffilFour}{Independent}
\newcommand{\CorrespondingEmail}{guanghan@fleet.so}

\title{Witness:  Discovery, Deciphering, and Epiphany \\ in Interactive Puzzle Environments}
\author{\AuthorOne$^{1}$\thanks{Corresponding author and project lead. Email: \texttt{\CorrespondingEmail}.}\quad
\AuthorTwo$^{2}$\quad \AuthorThree$^{3}$\quad \AuthorFour$^{4}$\quad \AuthorFive$^{1}$ \\
{\bfseries \AuthorSix$^{1}$\thanks{Equal contribution.}\quad \AuthorSeven$^{1}$\footnotemark[2]\quad \AuthorEight$^{1}$\footnotemark[2]\quad \AuthorNine$^{1}$} \\[4pt]
\parbox[t]{\dimexpr\textwidth-2\tabcolsep\relax}{\raggedright\mdseries $^{1}$\AffilOne \quad $^{2}$\AffilTwo \quad $^{3}$\AffilThree \quad $^{4}$\AffilFour}
}
\ppfinalcopy
\begin{document}
\maketitle
\begin{abstract}
Automated science needs agents that can work out the rules of an unfamiliar environment by interacting with it.
Interactive rule-discovery puzzles offer a controlled setting for studying this ability: an agent infers hidden rules through experimentation and uses what it has inferred to reach a stated goal.
We ask what limits current language models on these puzzles and whether reinforcement learning (RL) improves performance on rules held out from training.
To study both, we introduce WITNESS, a 2D grid-based puzzle environment with ground-truth ASCII observations and controlled access to rules.
An agentic pipeline generates games for WitnessGym, the RL training suite, and WitnessBench, comprising public validation and private test games.
The validation set separately tests new compositions of trained rule primitives and primitives absent from training.
Under a shared harness, the best of 18 frontier proprietary and open-weight models solves only 24\% of private test level slots, with scores sensitive to the observation interface and agent configuration.
Providing ground-truth rules raises Opus-5's validation RHAE-L5 (relative human action efficiency over the first five levels) from 59.9 to 97.8, whereas a 27B open-weight model gains only 2.1 points and remains limited even with the rules provided.
RL on WitnessGym raises the 27B model's private test RHAE-L5 from 2.1 to 5.4 and yields a mean gain of 4.1 points on four external discovery benchmarks.
Together, these results point to rule acquisition as a major difficulty for frontier models like Opus-5 while smaller models further struggle on rule-based execution, and indicate that RL on hidden-rule puzzles transfers to broader rules and real-world tasks beyond training.
Benchmark is available at: \url{https://witnessbench.ai}.

\end{abstract}
\section{Introduction}
%\vspace{-0.5em}
Discovering the rules governing an environment requires an agent to form hypotheses, choose informative actions, and revise its beliefs based on the resulting observations \citep{automation-of-science, where-science-starts}.
Interactive rule-discovery puzzles provide a controlled setting for studying this process: agents interpret observations, gather evidence about hidden rules, and use that evidence to guide their actions toward a goal.
However, these capabilities are entangled:
A low success rate does not by itself indicate whether state interpretation, rule discovery, or planning failed.
We therefore ask two questions: where current models struggle, and, beyond diagnosing limitations, whether reinforcement learning~(RL) improves performance on rules held out from training~\citep{discovering-faster-mm}.

%\vspace{-0.5em}
Addressing these questions requires two kinds of control: for diagnosis, control over the observation interface and over access to ground-truth rules; for transfer, training and evaluation sets separated by rule.
ARC-AGI-3 \citep{arcagi3}, the closest benchmark in spirit, evaluates rule discovery in visually presented games, where performance depends on both state interpretation and discovery.
Text-based benchmarks avoid the visual gap but drop something else: DiG-Bench \citep{DiG-bench} poses rule discovery over a single-line string with no spatial structure, and MazeBench \citep{Maze-bench} has spatial structure but rules are fixed and known to the model.
Other game benchmarks \citep{causal-game, puzzle-world, LMGame-bench, gg-bench, Enigmata} test reasoning over static patterns or over games whose rules are given, and failures on new puzzle variants~\citep{HardCoreLogic} raise concerns about generalization beyond those families.

\begin{figure}[t]
    \centering
    \includegraphics[width=1\textwidth]{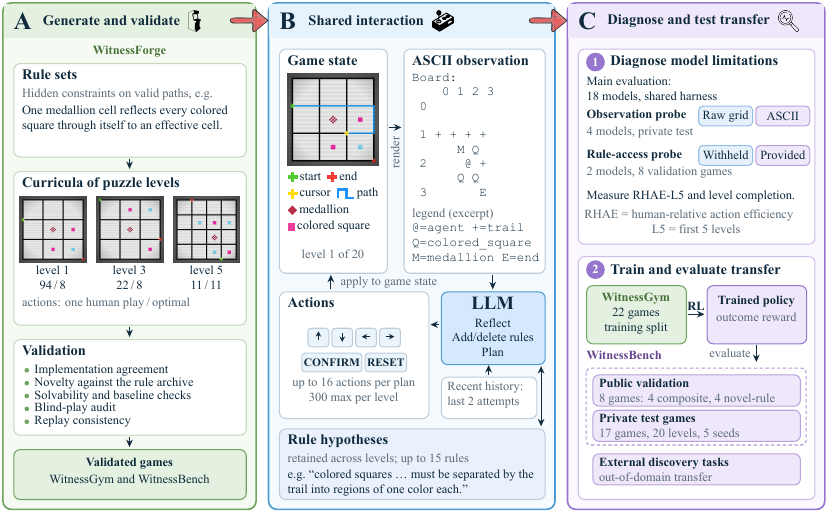}

    \vspace{-0.5em}
    \caption{Overview of \textbf{W}itness-\textbf{I}nspired \textbf{T}estbed of \textbf{N}ovel \textbf{E}nvironments with \textbf{S}ymbolic \textbf{S}tructure (WITNESS).}
    \label{fig:fig1_overview}
    \vspace{-2.0em}
\end{figure}

To provide both kinds of control, we introduce WITNESS (Figure~\ref{fig:fig1_overview}), an execution environment for 2D grid-based puzzles in the spirit of \textit{The Witness} \citep{thewitness2016}.
Each game asks the agent to draw a path from a start to a goal on a grid under hidden rules that constrain which paths are valid, and the engine exposes the board as a ground-truth ASCII representation, so the model parses spatial structure without visual recognition.
An agentic pipeline synthesizes new games with their own curricula of levels, from which we build WitnessGym for training and WitnessBench for evaluation, with public validation and private test sets.
To distinguish compositional transfer from transfer to new rule primitives, we organize the validation games by their relationship to the RL training rules: held-out compositions (new combinations of trained primitives) and held-out primitives (rules whose primitives do not appear in training).

On the first question, diagnosis in this environment under a shared harness indicates that acquiring the rules is a major difficulty for Opus-5, while substantial difficulty remains for Qwen3.8-27B even when the rules are provided.
Under a harness that standardizes observations, memory, and actions and permits no code execution, the best of 18 frontier proprietary and open-weight models, Fable-5, solves only 24\% of private test level slots.
To assess
score dependency on
observation interface, we replace ASCII rendering with the raw grid, which substantially reduces performance for several models while leaving Opus-5 largely unaffected.
In a separate rule-access probe, providing ground-truth rules raises Opus-5's validation relative human action efficiency over the first five levels \citep{arcagi3report} (RHAE-L5; higher is better) from 59.9 to 97.8, with all five evaluated levels solved in each of the 40 game runs.
For Qwen3.8-27B, the improvement is smaller, from 8.0 to 10.1 RHAE-L5.

The second question is whether RL on games with hidden rules improves performance on rules held out from training.
To answer this question, we train Qwen3.8-27B in WitnessGym using RL with outcome-based reward and evaluate its transfer to held-out games and external tasks.
The trained model improves from 2.1 to 5.4 RHAE-L5 on the private test set with thinking off, and achieves higher validation performance on both held-out compositions and held-out primitives.
Beyond the suite, the trained model achieves a mean gain of 4.1 percentage points (95\% CI 2.1 to 6.2) on four external benchmarks of discovery-related reasoning \citep{CipherBank, PhysGym, wilt, failing-to-falsify}, suggesting that part of what is learned is not specific to these puzzles.
We observe no improvement for Qwen3.5-9B under the same training setup, and credit assignment keyed to surviving rule hypotheses did not yield additional gains over broadcasting episode advantages.

Taken together, these diagnostic and training results support three contributions:
\begin{itemize}[itemsep=2pt, topsep=0pt, parsep=0pt,leftmargin=*]
\item WITNESS, an environment and generation pipeline for grid-based rule-discovery puzzles with a text interface, from which we release WitnessGym for training and WitnessBench for evaluation, with validation rules held out from RL training as held-out compositions and held-out primitives.
\item An evaluation of 18 models under a shared harness, with controlled observation and rule-access probes indicating that acquiring the rules is a major difficulty for Opus-5 and that difficulty persists for Qwen3.8-27B when the rules are provided.
\item Evidence that RL with outcome-based reward improves performance on the private test set, on held-out compositions and held-out primitives, and on four external benchmarks, together with an empirical comparison in which credit assignment keyed to surviving hypotheses did not yield additional gains over broadcasting episode advantages.
\end{itemize}

\vspace{-0.5em}

\section{Benchmark}
\label{section:benchmark}
\vspace{-0.5em}

\begin{figure}[t]
    \centering
    \includegraphics[width=\textwidth]{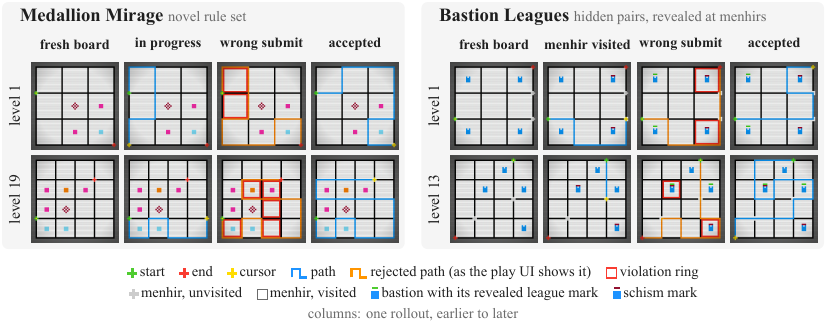}
    \vspace{-2em}
    \caption{One level of a WITNESS game as the engine renders it and as the model observes it: the ASCII board with its legend, the current path, and the feedback shown after a rejected submission.}
    \label{fig:fig2_gameplay}
    \vspace{-1em}
\end{figure}

WITNESS comprises a game engine, a suite of games with WitnessBench for evaluation and WitnessGym for training, a pipeline that generates and validates new games, a shared evaluation harness, and an action-efficiency metric.

\vspace{-0.5em}
\subsection{Game Engine and Mechanics}
\label{section:mechanics}

The engine is built on the official ARC-AGI-3 SDK \citep{arc_agi_toolkit} and renders each frame as a $64\times64$ pixel image.
It also emits a ground-truth ASCII representation of the board, which serves as the model's observation (Figure~\ref{fig:fig2_gameplay}).
The board is an $M\times N$ grid on which the agent draws a path from a start cell to a goal cell; a level is solved when the submitted path satisfies every rule of the game.
The action space consists of four directional moves, SUBMIT, and RESET.
The semantics of SUBMIT and RESET are stated to the model, but the mapping from the remaining action IDs to directions has to be inferred through interaction.

Game rules constrain which paths are valid, and discovering them is left to the agent.
Classic games use publicly documented primitives from \textit{The Witness}, while generated games use rules constructed and validated by WitnessForge (Section~\ref{section:crafting-games}).
The classic primitives may have appeared in model pretraining; throughout the paper, held-out status refers only to the RL training split.

The engine provides two kinds of feedback.
An invalid move, such as crossing a wall or a broken edge, is refused and leaves the frame unchanged.
A submitted path that violates a rule is rejected, with the path and violating cells highlighted in both the image and the ASCII representation.
These highlights depend on the submitted path, allowing different attempts to reveal different evidence about the rule.
Some games, such as \textit{Bastion Leagues} in Figure \ref{fig:fig2_gameplay}, additionally require interactions that reveal hints: the agent has to discover how to elicit these hints before using them to plan a solution.

\subsection{Game Suite: WitnessBench and WitnessGym}
\label{section:games}

WitnessBench is the benchmark: 8 public validation games and 17 private test games, on which every model comparison in this paper is reported.
WitnessGym is its training counterpart: 22 public games that share the engine and harness with the benchmark but none of its rules.
Every private test game consists of 20 levels: the first five use small grids and introduce the rules; remaining levels use larger grids on which the difficulty shifts to planning and executing valid paths under rules already learned.

The training set contains 8 classic games, each based on one primitive rule of \textit{The Witness}, and 14 generated games.
The validation set tests generalization at two levels: 4 classic games combine training primitives in new ways (held-out \emph{compositions}), and 4 generated games use primitives absent from training (held-out \emph{primitives}).
The composition split deliberately reuses training primitives, while the primitive split tests transfer beyond them.
The 17 test games are generated with rules not used in training and are kept private; the training and validation games are public, and the generation pipeline~(Section~\ref{section:crafting-games}) supports periodic refresh of private set like LiveCodeBench \citep{jain2025livecodebench}.

\vspace{-0.5em}
\subsection{Game Generation and Validation}
\label{section:crafting-games}
\vspace{-0.5em}

WitnessForge constructs and validates the generated games in both suites.
Given a rule concept in natural language, it produces a packaged game containing a rule implementation, a curriculum of levels, and metadata, extensible without manually implementing rules or constructing levels.
Its foundation is a path-enumeration kernel written in Rust: given a board, it enumerates every simple path from a start node to the goal node, and a predicate layer applies the rule-checking functions to return the exact set of valid solutions.
The pipeline has two stages: rule verification followed by level design and audit; Algorithm~\ref{alg:game_pipeline} in Appendix~\ref{section:pipeline} gives the full procedure and the filtering thresholds.

\noindent\textbf{Rule verification.}
A candidate rule is admitted after passing a series of programmatic checks; besides cheap static checks,
 three semantic checks carry the argument.
First, two implementations are generated independently from the same specification and compared over enumerated path spaces on test boards; counterexamples guide revision for a bounded number of rounds, and candidates whose implementations still disagree are rejected.
Second, a solution-space ablation removes the rule and checks that the valid-solution set strictly expands, showing that the rule excludes otherwise valid paths on the tested boards.
Third, a novelty check screens for distinctness by comparing each candidate against the existing rule archive.
Rules that reduce strictly to path-length checks are pruned as degenerate.
These checks provide evidence of distinctness relative to the tested boards and archived rules, rather than guaranteeing novelty relative to pretraining data.

\noindent\textbf{Level design and audit.}
Levels are designed to distinguish the intended rule from a predefined set of incorrect candidates through differences in their valid-solution sets, with each incorrect candidate refuted by at least one level's stored solution.
For each admitted rule, the pipeline generates candidate levels and retains those that pass structural validation, whose shortest solution meets a minimum action count, and on which removing the rule strictly enlarges the valid-solution set.
It then filters out levels solved by the tested rule-agnostic heuristics or greedy baselines, or whose exact random-play success probability exceeds a fixed threshold.
The retained levels are annotated with the candidates they refute and ordered into a curriculum that introduces the rule and eliminates remaining alternatives, generating additional levels when the pool lacks a discriminating one.
The assembled game then undergoes three final checks.
A held-out language model plays it blind to test whether its rule can be recovered from interaction alone; an independent audit re-implements all filters from scratch to probe levels with fresh strategies and seeds; and each stored solution is replayed against the packaged game to verify solvability.
Across the released games, 5 of 42 candidate rules were rejected, most often for their unlearnability across models; 97\% candidate levels were rejected by the programmatic filters, as random play could solve them or they admitted too many solutions, preserving 760 levels.

\vspace{-0.5em}
\subsection{Evaluation Harness}
\vspace{-0.5em}
\label{section:harness}

All models interact with the engine through the same harness, which standardizes observations, memory support, and available actions.
Models reason in text and interact with the game through six actions, without code execution or external tools.
At each turn, the harness presents the current board as ASCII text with a symbol legend.
Alongside the board, it provides the path drawn so far, the previous action and its effect, and a record of recent attempts on the current level and their outcomes.
To support reasoning across attempts and levels, the observation also includes a rule memory containing hypotheses recorded by the model.
This memory persists within each game and is cleared between games.
Given this observation, the model responds with free-form analysis, updates to its rule memory (add, delete, or keep), and an optional sequence of up to 16 actions.
The engine executes the sequence before returning the next observation, stopping early if the level is solved.
If the model returns no sequence, the harness instead executes one random action, which counts toward the action budget like any other.
Interaction continues until the final level is solved or a level remains unsolved after its budget of 300 actions per level or reaching 1000 LLM calls per game; an unsolved level ends the run rather than being skipped.
Complete prompts are provided in Appendix~\ref{section:prompts}.

\subsection{Evaluation Metric and Protocol}
\label{section:metric}

The evaluation metric is Relative Human Action Efficiency (RHAE), taken from ARC-AGI-3~\citep{arcagi3report}.
A solved level is scored by the agent's action count $S^\ell_{\mathrm{agent}}$ against a reference count $S^\ell_{\mathrm{ref}}$, capped by $1.15$; an unsolved or unreached level scores zero:
\begin{equation}
\small
\mathrm{score}_\ell =
\begin{cases}
\min\!\Big( \big(S^\ell_{\mathrm{ref}}/S^\ell_{\mathrm{agent}}\big)^{2},\; 1.15 \Big) & \text{if level $\ell$ is solved,}\\[2pt]
0 & \text{otherwise.}
\end{cases}
\label{eq:arc-score-level}
\vspace{-0.5em}
\end{equation}
The reference count is the mean action count across successful plays by team members who had no access to the rule specification, counted on the same basis as the agent's actions, including failed attempts and resets.
All levels in validation and test set games have successful human play data.

For a game $g$ evaluated over its first $L$ levels, the game score weights each level by its index $\ell$ and is capped at the weighted completion fraction:
\begin{equation}
\small
\mathrm{Score}_g \;=\; 100\cdot \min\!\left(
  \frac{\sum_{\ell=1}^{L} \ell\cdot\mathrm{score}_\ell}{\sum_{\ell=1}^{L} \ell},\;
  \frac{\sum_{\ell=1}^{L} \ell\cdot\mathds{1}[\text{level $\ell$ solved}]}{\sum_{\ell=1}^{L} \ell}
\right).
\label{equation:arc-score-game}
\end{equation}
Because each level score is at most $1.15$, the first term can exceed the second only when the weighted mean efficiency exceeds one; the minimum therefore allows above-reference efficiency on one level to offset below-reference efficiency on another, but never to offset an unsolved level.
We report RHAE-L5 over the first five levels ($L = 5$) as the headline metric, emphasizing introductory performance, and RHAE-L20 over all twenty levels ($L = 20$) to measure full curriculum performance.
The game engine stores the optimal step $S^\ell_{\mathrm{optim}}$ for each level $\ell$.
Objective action efficiency (where $S^\ell_{\mathrm{ref}} = S^\ell_{\mathrm{optim}}$) and fully uncapped RHAE scores are also reported for reference and analysis.

For each seed, the dataset score is the mean game score across the split (17 test or 8 validation games).
We report the mean and standard deviation across $M = 5$ test seeds and $M = 3$ validation seeds (except Table \ref{tab:gt-rule-card-probe} in Appendix \ref{section:gt-rule-ablation}, using 5 seeds for GT-rule ablation), giving 85 test runs per model.

\section{Model Evaluation: What Limits Current Models}
\label{section:model-performance}
\vspace{-0.5em}

This section addresses the first question of the introduction, what limits current models, in three steps.
Section~\ref{section:overall-performance} reports the overall performance of 18 models on the private test set under the shared harness.
Section~\ref{section:harness-ablations} compares this harness with alternative observation and harness settings to measure how much the scores depend on them.
Section~\ref{section:gt-rules} measures the effect of providing the ground-truth rules on the validation set, to isolate the rule acquisition difficulty.

\vspace{-0.5em}
\subsection{Overall Performance}
\vspace{-0.5em}
\label{section:overall-performance}
Table~\ref{tab:model_evals} reports 18 proprietary and open-weight models on the 17 private test games, with 5 seeds per game, giving 85 runs and 1,700 level slots per model.
All models use the harness of Section~\ref{section:harness} and the scoring rule of Section~\ref{section:metric}; each model runs at its highest reasoning setting, with per-model configurations and full per-model statistics in appendices.
Two findings stand out from the table.

First, under the shared harness, all evaluated models solve only a minority of the available level slots, and complete runs are rare.
The model with the highest mean RHAE-L5, Fable-5, reaches $36.4 \pm 3.7$, solves 414 of 1,700 level slots (4.87 levels per run), and completes 2 of 85 runs.
The lowest-scoring model, Qwen3.5-9B, reaches $1.0 \pm 0.8$, solves 44 level slots (0.52 levels per run), and completes no run.
Twelve of the 18 models complete no run.
Qwen3.8-27B, the starting checkpoint used for training in Section~\ref{section:training-gym}, reaches an RHAE-L5 of 6.4 at its highest reasoning setting; Section~\ref{section:training-gym} evaluates this model with thinking off, so its numbers there are not comparable to this table.

Second, the ranking depends on the evaluation window.
GPT-6-Astra-Pro has the highest RHAE-L20 at $12.8 \pm 1.1$ and completes 10 of 85 runs, whereas Fable-5 scores 8.3 on RHAE-L20 with 2 completed runs.
Performance on the first five levels and performance over the full curriculum therefore give different orderings, and a higher early score does not imply more completed runs.
We report both windows; Sections~\ref{section:harness-ablations} and \ref{section:gt-rules} examine what the scores depend on.

\definecolor{rhaefill}{HTML}{27F5BB}
\definecolor{rhaefillL}{HTML}{F5CBA7}
\definecolor{rhaetrack}{HTML}{EEF0F3}
\definecolor{rhaeink}{HTML}{15202B}
\ifdefined\rhaebarwidth\else\newlength{\rhaebarwidth}\newlength{\rhaebarfill}\fi
\setlength{\rhaebarwidth}{58pt}
\providecommand{\rhaebarmax}{40}
\providecommand{\rhaebarLmax}{15}
\providecommand{\rhaebarcore}[4]{\setlength{\rhaebarfill}{\dimexpr #1\rhaebarwidth/#3\relax}
  \ifdim\rhaebarfill>\rhaebarwidth \setlength{\rhaebarfill}{\rhaebarwidth}\fi\makebox[\rhaebarwidth][l]{\rlap{\textcolor{rhaetrack}{\rule[-2.1pt]{\rhaebarwidth}{10pt}}}\rlap{\textcolor{#4}{\rule[-2.1pt]{\rhaebarfill}{10pt}}}\makebox[\rhaebarwidth][r]{\textcolor{rhaeink}{#2}\hspace{2.5pt}}}}
\providecommand{\rhaebar}[2]{\rhaebarcore{#1}{#2}{\rhaebarmax}{rhaefill}}
\providecommand{\rhaebarL}[2]{\rhaebarcore{#1}{#2}{\rhaebarLmax}{rhaefillL}}

\begin{table}[t]
\centering\scriptsize\setlength{\tabcolsep}{2.5pt}\renewcommand{\arraystretch}{1.1}
\caption{
\textbf{Model Evals using our canonical harness} (unseen games w/ novel rules, 20 lvls/game, 5 eval seeds).
RHAE-L5 / RHAE-L20: RHAE computed over the first 5 levels (the benchmark number) and over all 20 levels.
The first 5 levels are four rule-teaching levels with curriculum design plus a non-teaching testing level; the remaining levels are more complex levels of larger grid or harder.
RHAE-uncap: the human-baseline score without the two caps.
OAE (objective action efficiency): the same level-weighted efficiency score against the optimal action count.
Deepest level: the most levels completed in any single rollout.
Evaluated in \textbf{max-effort} where the endpoint exposed effort tiers.
}
\label{tab:model_evals}
\vspace{-1em}
\begin{tabular}{@{}l c c c c c c c c@{}}
\toprule
Model & \begin{tabular}[c]{@{}c@{}}RHAE-L5\\mean $\pm$ std\end{tabular} & \begin{tabular}[c]{@{}c@{}}Beaten\\levels\end{tabular} & \begin{tabular}[c]{@{}c@{}}Avg levels\\ per game\end{tabular} & \begin{tabular}[c]{@{}c@{}}OAE\\L5\,/\,L20\end{tabular} & \begin{tabular}[c]{@{}c@{}}RHAE-uncap\\L5\,/\,L20\end{tabular} & \begin{tabular}[c]{@{}c@{}}Deepest\\level\end{tabular} & \begin{tabular}[c]{@{}c@{}}Beaten\\games\end{tabular} & \begin{tabular}[c]{@{}c@{}}RHAE-L20\\mean $\pm$ std\end{tabular} \\
\midrule
Fable-5 & \rhaebar{36.4}{36.4 $\pm$ 3.7} & 414/1700(24\%) & 4.87/20 & 20.9\,/\,5.4 & 156.3\,/\,34.3 & 20/20 & 2/85 & \rhaebarL{8.3}{8.3 $\pm$ 2.7} \\
Opus-5 & \rhaebar{30.6}{30.6 $\pm$ 2.0} & 405/1700(24\%) & 4.76/20 & 21.2\,/\,6.6 & 210.9\,/\,38.6 & 20/20 & 2/85 & \rhaebarL{9.4}{9.4 $\pm$ 2.2} \\
GPT-6-Astra-Pro & \rhaebar{29.8}{29.8 $\pm$ 2.0} & 399/1700(23\%) & 4.69/20 & 19.5\,/\,9.1 & 224.1\,/\,59.5 & 20/20 & 10/85 & \rhaebarL{12.8}{12.8 $\pm$ 1.1} \\
GPT-6-Astra & \rhaebar{27.6}{27.6 $\pm$ 0.9} & 389/1700(23\%) & 4.58/20 & 17.6\,/\,7.7 & 182.7\,/\,50.0 & 20/20 & 10/85 & \rhaebarL{11.3}{11.3 $\pm$ 0.9} \\
Kimi-K3 & \rhaebar{24.6}{24.6 $\pm$ 4.0} & 249/1700(15\%) & 2.93/20 & 14.9\,/\,1.8 & 119.2\,/\,15.2 & 20/20 & 1/85 & \rhaebarL{3.1}{3.1 $\pm$ 1.6} \\
GPT-5.6-Sol & \rhaebar{23.0}{23.0 $\pm$ 3.1} & 262/1700(15\%) & 3.08/20 & 13.8\,/\,2.1 & 92.7\,/\,10.8 & 14/20 & 0/85 & \rhaebarL{3.4}{3.4 $\pm$ 1.0} \\
Opus-4.8 & \rhaebar{22.8}{22.8 $\pm$ 4.3} & 203/1700(12\%) & 2.39/20 & 14.4\,/\,1.1 & 106.2\,/\,7.8 & 9/20 & 0/85 & \rhaebarL{1.8}{1.8 $\pm$ 0.3} \\
Muse-Spark-1.2 & \rhaebar{14.9}{14.9 $\pm$ 0.4} & 194/1700(11\%) & 2.28/20 & 8.2\,/\,0.7 & 28.2\,/\,2.3 & 11/20 & 0/85 & \rhaebarL{1.2}{1.2 $\pm$ 0.1} \\
Grok-4.6 & \rhaebar{12.9}{12.9 $\pm$ 2.6} & 240/1700(14\%) & 2.82/20 & 5.8\,/\,0.7 & 43.2\,/\,4.8 & 20/20 & 1/85 & \rhaebarL{1.6}{1.6 $\pm$ 1.1} \\
DeepSeek V4 Pro & \rhaebar{7.1}{7.1 $\pm$ 2.3} & 132/1700(8\%) & 1.55/20 & 4.0\,/\,0.3 & 47.8\,/\,3.4 & 5/20 & 0/85 & \rhaebarL{0.5}{0.5 $\pm$ 0.2} \\
Qwen-3.8-Max & \rhaebar{6.9}{6.9 $\pm$ 1.0} & 132/1700(8\%) & 1.55/20 & 4.0\,/\,0.3 & 26.8\,/\,1.9 & 5/20 & 0/85 & \rhaebarL{0.5}{0.5 $\pm$ 0.1} \\
Qwen3.8-27B & \rhaebar{6.4}{6.4 $\pm$ 1.2} & 112/1700(7\%) & 1.32/20 & 3.9\,/\,0.3 & 16.8\,/\,1.2 & 5/20 & 0/85 & \rhaebarL{0.5}{0.5 $\pm$ 0.1} \\
GLM-5.2-Max & \rhaebar{5.6}{5.6 $\pm$ 0.8} & 109/1700(6\%) & 1.28/20 & 2.8\,/\,0.2 & 20.8\,/\,1.5 & 5/20 & 0/85 & \rhaebarL{0.4}{0.4 $\pm$ 0.1} \\
GLM-5.3 & \rhaebar{4.7}{4.7 $\pm$ 2.6} & 90/1700(5\%) & 1.06/20 & 2.7\,/\,0.2 & 9.1\,/\,0.7 & 5/20 & 0/85 & \rhaebarL{0.3}{0.3 $\pm$ 0.2} \\
Qwen3.5-35B-A3B & \rhaebar{2.8}{2.8 $\pm$ 1.3} & 58/1700(3\%) & 0.68/20 & 1.7\,/\,0.1 & 3.6\,/\,0.3 & 4/20 & 0/85 & \rhaebarL{0.2}{0.2 $\pm$ 0.1} \\
Glim-30 & \rhaebar{2.7}{2.7 $\pm$ 0.7} & 57/1700(3\%) & 0.67/20 & 1.9\,/\,0.1 & 5.6\,/\,0.4 & 4/20 & 0/85 & \rhaebarL{0.2}{0.2 $\pm$ 0.1} \\
Dots-3-Note-Preview & \rhaebar{2.6}{2.6 $\pm$ 1.7} & 54/1700(3\%) & 0.64/20 & 0.9\,/\,0.1 & 10.6\,/\,0.8 & 5/20 & 0/85 & \rhaebarL{0.2}{0.2 $\pm$ 0.1} \\
Qwen3.5-9B & \rhaebar{1.0}{1.0 $\pm$ 0.8} & 44/1700(3\%) & 0.52/20 & 0.6\,/\,0.0 & 1.4\,/\,0.1 & 3/20 & 0/85 & \rhaebarL{0.1}{0.1 $\pm$ 0.1} \\
\bottomrule
\end{tabular}
\end{table}

\begin{table}[!t]
\centering
\scriptsize
\setlength{\tabcolsep}{4pt}
\caption{
\textbf{Harness ablation on the 17 unseen games} (5 seeds): how much of each model's RHAE-L5 score depends on its harness.
\emph{Our harness}: our full harness.
\emph{w/o perception}: same harness without ground-truth ASCII board ( $64{\times}64$ pixel input).
\emph{Minimal harness}: a minimalist loop, raw pixel input.
\emph{Prime-Agent}:
A self-improving RLM Harness (95\% on ARC-AGI-3 public games).
\emph{Model-attended solving} and \emph{Verbal-only solving}: exclude cases -- no model participation in level solving; winning moves computed by code.
Details in Appendix~\ref{section:oss-harness-mechanism}.
$^{*}$These models with Prime-Agent caused frequent refusals from Claude's content filter; we report 2 seeds to show a spread and repeat no further.
}
\label{table:harness_ablation}
\resizebox{0.9\linewidth}{!}{
\begin{tabular}{@{\extracolsep{\fill}}lcccc@{}}
\toprule
Harness                           & Opus-5 & Opus-4.8 & Kimi-K3 & Qwen3.8-27B \\
\midrule
\addlinespace[2pt]
Our harness                       & 30.58 $\pm$ 2.00 & 22.77 $\pm$ 4.27 & 24.62 $\pm$ 4.02 & 6.42 $\pm$ 1.19 \\
Our harness, w/o perception       & 30.29 $\pm$ 4.92 & 10.95 $\pm$ 2.80 & 8.71 $\pm$ 2.91 & 1.03 $\pm$ 0.30 \\
Minimal harness (ARC-AGI-3 alike) & 11.33 $\pm$ 3.02 & 3.31 $\pm$ 1.11 & 0.00 $\pm$ 0.00 & 0.01 $\pm$ 0.03 \\
\addlinespace[1pt]
\hdashline[1pt/1.5pt]
\addlinespace[3pt]
OSS harness: Prime-Agent          & 68.11 $\pm$ 6.33$^{*}$ & 59.88 $\pm$ 0.75$^{*}$ & 47.99 $\pm$ 3.96 & 26.60 $\pm$ 6.39 \\
\quad model-attended solving      & 55.29 $\pm$ 6.68 & 37.24 $\pm$ 4.66 & 42.22 $\pm$ 4.53 & 23.53 $\pm$ 3.87 \\
\quad verbal-only solving         & 4.37 $\pm$ 1.45 & 0.14 $\pm$ 0.05 & 9.20 $\pm$ 4.38 & 5.86 $\pm$ 3.90 \\
\bottomrule
\end{tabular}
}
\end{table}

\subsection{Observation and Harness Comparisons}
\label{section:harness-ablations}
Table~\ref{tab:model_evals} shows scores of a model together with a harness, so we measure how much they depend on the harness in two directions, removing components from it and adding tools beyond it in Table~\ref{table:harness_ablation}.

\noindent\textbf{Removing harness components.}
Replacing the ASCII board with the raw $64{\times}64$ grid changes only the observation.
RHAE-L5 then falls from 30.58 to 30.29 for Opus-5, from 22.77 to 10.95 for Opus-4.8, from 24.62 to 8.71 for Kimi-K3, and from 6.42 to 1.03 for Qwen3.8-27B.
The size of the drop differs widely across models, and the ordering of the mean scores of Opus-4.8 and Kimi-K3 reverses between the two settings, so the observation representation affects both the scores and the observed comparisons between models.
The minimal harness goes further and removes the observation, the memory, and the interaction support at once.
Under it, RHAE-L5 is 11.33 for Opus-5, 3.31 for Opus-4.8, and at most 0.01 for the other two models.
Because several components are removed together, this result shows the combined effect of multiple harness components.

\noindent\textbf{Expanded tool setting.}
Prime-Agent adds code execution around the model, so it evaluates the same games under an expanded tool setting, and the comparison measures performance under different agent configurations rather than isolating the effect of code execution alone.
It achieves higher scores for all four models (68.11, 59.88, 47.99, and 26.60 on RHAE-L5), showing that measured performance depends strongly on the surrounding agent configuration.
To audit how the levels were solved, we rescore the same runs post hoc by trajectory behaviour: one rescoring excludes levels on which code submitted three or more candidate paths in one turn and let the engine verdict select the answer, and the other keeps only levels whose winning moves the model produced without code.
With typed moves only, the scores fall to 4.37, 0.14, 9.20, and 5.86, so most credited levels involved programmatic search or solution generation.
These rescored values describe the runs as they happened and are not estimates of what the models would score without code, because later levels may build on earlier code-based solutions and code that fits candidate rules may itself take part in hypothesis construction.
More details and example trajectories are given in Appendix~\ref{section:oss-harness-mechanism}.

\vspace{-0.5em}
\subsection{Effect of Providing Ground-Truth Rules}
\label{section:gt-rules}
\vspace{-0.5em}
\begin{table}[t]
\centering
\scriptsize
\caption{\textbf{Ground-truth rule cards as a prompt-time probe}.
A ground-truth rule card is a hand-written natural-language statement of a game's true rules (win condition and mechanics, transcribed from the game engine), which we supply in the system prompt at evaluation time only, as an oracle probe that separates a policy's failure to discover the rules from its failure to act on them.
Eight held-out validation games, 5 evaluation seeds, 8192-token output cap, thinking off.
}
\vspace{-1em}
\label{tab:gt-rule-card-main}
\resizebox{\textwidth}{!}{\begin{tabular}{llccccc}\toprule
Policy & Extra prompt & RHAE-L5 & $\Delta$RHAE-L5 & Games cleared & Efficiency & Plan-less calls \\
\midrule
Qwen3.8-27B & none & 8.0 $\pm$ 5.0 & --  & 5/40 & 0.39 & 1\% \\
 & generic text & 9.1 $\pm$ 2.5 & +1.1  & 6/40 & 0.44 & 0\% \\
 & GT rule card & 10.1 $\pm$ 2.7 & +2.1  & 6/40 & 0.48 & 1\% \\
\midrule
Opus 5 (frontier) & none & 59.9 $\pm$ 11.4 & --  & 30/40 & 0.74 & 3\% \\
 & generic text & 63.6 $\pm$ 3.7 & +3.7 &  34/40 & 0.72 & 4\% \\
 & GT rule card & 97.8 $\pm$ 0.7 & +37.8 & 40/40 & 0.99 & 0\% \\
\bottomrule
\end{tabular}
}
\vspace{-2em}
\end{table}

Providing ground-truth rules probes how much performance improves when the model no longer needs to infer the stated game mechanics from interaction.
For each of the eight validation games we write a rule card, a natural-language statement of its win condition and mechanics transcribed from the engine, including the action mapping, and place it in the system prompt at evaluation time.
As a control, a generic-text prompt of matched length gives the same action mapping together with general guidance on observation, hypothesis testing, and efficiency, but states no rule of the game.
Both prompts include the same action mapping and are matched in length, while differing in their task-specific content; details in Appendix~\ref{section:gt-rule-ablation}.
The probe uses 5 seeds, thinking off, an 8,192-token output cap, and the first five levels of each game, a configuration different from Table~\ref{tab:model_evals}, so scores are not comparable across tables.
Table~\ref{tab:gt-rule-card-main} reports Opus-5 and the Qwen3.8-27B starting checkpoint; Appendix~\ref{section:gt-rule-ablation} reports the full probe, including the RL-trained checkpoints discussed in Section~\ref{section:training-gym}.

The two models respond to the rule card very differently.
Relative to the length-matched generic prompt, the rule card raises Opus-5 by 34.1 points to 97.8 RHAE-L5 with every evaluated level solved in all 40 game runs, so explicit rule access is sufficient to close the gap to the ceiling in this probe; this indicates that acquiring the rules is a major difficulty for this model on these games.
For Qwen3.8-27B, the rule card yields an observed mean difference of 1.0 points over the generic prompt, to 10.1, with 6 of 40 game runs clearing all evaluated levels in both conditions.
Substantial difficulty therefore still remains for this model when the rules are provided, indicating limitations beyond acquiring the rule description.
These results for the two models describe them as they are; whether RL on games with hidden rules improves performance on rules held out from training, and whether any improvement extends beyond the suite, is
the subject of Section~\ref{section:training-gym}.

%\vspace{-0.5em}

\section{RL Training: Does It Transfer beyond Training Rules}
\label{section:training-gym}
%\vspace{-1em}

This section addresses the second question of the introduction, whether RL on games with hidden rules improves performance on rules held out from training.
We describe the training setup, report learning and transfer within the suite on the validation and private test sets, compare credit-assignment schemes, and report transfer to external benchmarks.

\begin{figure}[t]
    \centering
    \includegraphics[width=\textwidth]{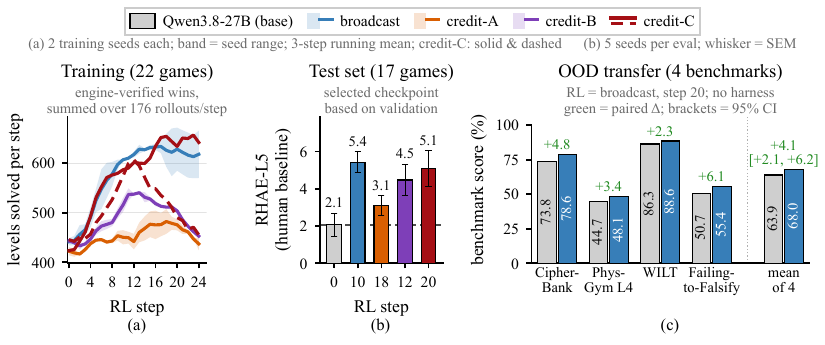}
    \vspace{-2em}
    \caption{(a)~RL training dynamics; and (b,c)~generalization to test set (unseen rules) and external discovery tasks. }
    \label{fig:fig_compact_rl_ood}
    \vspace{-2em}
\end{figure}

\noindent\textbf{Training setup.}
We train Qwen3.5-9B and Qwen3.8-27B on the 22 training games of Section~\ref{section:games} with the Miles framework \citep{miles2026}.
Each training episode plays one game from its first level under the harness of Section~\ref{section:harness}, using game levels capped at first ten, and ends when all of them are solved, after 300 LLM calls, or when a level remains unsolved after its 300-action budget.
The reward is outcome-based: $+1$ per solved level, $+0.5$ for solving all levels of the game, and $-0.005$
per LLM call to encourage action efficiency.
The baseline algorithm is GRPO \citep{GRPO} with group centering by game, batch-level whitening, and asymmetric clipping, and it broadcasts each episode's advantage to every LLM call of that episode; Appendix~\ref{section:rl-settings} gives the objective and hyperparameters.
Each iteration samples 8 episodes for each games, and training runs for 30 iterations.
Thinking is disabled during training so that the model's reasoning goes into the \texttt{<meta>} field required by the harness and rollouts stay short; all numbers in this section are therefore evaluated with thinking off and are not comparable to Table~\ref{tab:model_evals}.
Checkpoints are selected on the 8 validation games with 3 seeds, and the selected checkpoint is evaluated on the 17 private test games with 5 seeds.

\noindent\textbf{Learning and transfer within the suite.}
Figure~\ref{fig:fig3_scale_9b_vs_27b_rkae} in Appendix \ref{section:rl-settings} shows training reward and validation scores for both models under identical settings.
For Qwen3.8-27B, training reward rises over iterations, and validation RHAE-L5 rises on both held-out compositions and held-out primitives.
The two validation groups start at different levels, with held-out primitives near zero, and reach similar values by the end of training.
For Qwen3.5-9B, neither training reward nor validation score improves under the same settings.
On the private test set, the selected Qwen3.8-27B checkpoint raises RHAE-L5 from 2.1 to 5.4 and the number of solved level slots from 67 to 122 out of 1,700.
These gains are measured on rules absent from training, so they reflect transfer beyond the training games; what the model has learned that transfers is examined on the external benchmarks below.

\noindent\textbf{Credit assignment.}
Outcome-only reward leaves open how credit is distributed over the LLM calls of an episode, and we compare four schemes on Qwen3.8-27B.
Broadcast assigns the episode advantage to every call.
Credit-A groups calls by attempt within each level and computes credit per group, which makes the signal sparser.
Credit-B increases the weight of the last successful attempt within each level.
Credit-C additionally increases the weight of earlier attempts that proposed a rule hypothesis surviving to the end of a successful episode.
Figure~\ref{fig:fig_compact_rl_ood} and Appendix \ref{section:rl-settings} show that within the family, credit-A is below credit-B, which is below credit-C, and that credit-C is on par with broadcast.
The two credit-C seeds diverge early in training, which we attribute to the noisiness of survival as a proxy for hypothesis correctness: a hypothesis can survive an episode without being correct, and the resulting credit is then misdirected.
We therefore use broadcast for the remaining experiments.
The behavioral study along the RL training is shown in Figure~\ref{fig:fig5_behaviour_9b_vs_27b_survived}~(in Appendix~\ref{section:rl-settings}).

\noindent\textbf{Transfer to external benchmarks.}
To test whether the gains extend beyond grid puzzles, we evaluate the base model and the RL model (broadcast) on four external benchmarks of rule inference and hypothesis testing, all with thinking on and a 32k-token cap.
CipherBank \citep{CipherBank} measures exact-match accuracy on the 786 items of its author-invented ciphers.
PhysGym \citep{PhysGym} measures the success rate on its 97 level-4 problems, which give the least prior description of the physical system.
WILT \citep{wilt} measures elim@5, the fraction of predefined hypotheses eliminated within the first five experiments.
Failing-to-Falsify \citep{failing-to-falsify} measures the incompatibility fraction between a model's stated hypothesis and the tests it chooses, where a higher value indicates more falsification-seeking behavior.
At iteration 20, the broadcast checkpoint improves the mean over the four benchmarks by 4.1 points (95\% CI 2.1 to 6.2), with per-benchmark intervals excluding zero for CipherBank (+4.8) and WILT (+2.3), a lower bound at zero for PhysGym (+3.4), and an interval including zero for Failing-to-Falsify (+6.1).
These results indicate that part of what is learned on WitnessGym transfers to rule inference outside the suite, with the clearest gains on cipher decryption and experiment design.
Table~\ref{tab:dose_response_2x4} in Appendix \ref{section:appendix-ood-generalization} reports more detailed scores.

%\vspace{-0.5em}

\section{Related Work}
%\vspace{-0.5em}
\label{section:related-work}

\noindent\textbf{Interactive rule discovery and puzzle benchmarks.}
Puzzle and game benchmarks differ in what the agent observes and in how training and evaluation are divided; Table~\ref{tab:benchmark_comparison} compares these properties.
ARC-AGI-3 \citep{arcagi3} presents interactive games as integer grid frames whose rules and goals are not given, so a score reflects both the interpretation of the frames and the discovery of the rules.
Other listed benchmarks avoid visual recognition such as DiG-Bench~\citep{DiG-bench} and MazeBench~\citep{Maze-bench}.
Some benchmarks~\citep{gg-bench,Enigmata,causal-game, puzzle-world} state the rules in the prompt, whereas lmgame-Bench~\citep{LMGame-bench} uses published game titles whose mechanics may have appeared in pretraining data.
HardcoreLogic~\citep{HardCoreLogic} transforms familiar puzzle families into long-tail variants and reports large performance drops on the variants.
Relative to these designs, WITNESS exposes an entity-level two-dimensional board, hides the rules, and organizes validation games by their relationship to the RL training rules.
This combination supports controlled observation and rule-access probes, together with evaluation of transfer across specified rule splits.

\begin{table}[!t]
\centering\small\setlength{\tabcolsep}{4pt}
\caption{Comparison of related benchmarks. Hidden rules indicates interactive inference of rules not supplied to the agent. Split identifies the unit held out between training and evaluation in the reported experiments; evaluation only indicates that no training experiment is reported for evaluated models.}
\label{tab:benchmark_comparison}
\vspace{-0.5em}
\resizebox{\linewidth}{!}{
\begin{tabular}{l c l l}
\toprule
Benchmark & \makecell{Hidden \\ rules} & Observation & Training--evaluation split \\
\midrule
ARC-AGI-3 \citep{arcagi3}      & \gcmark & Pixel grid & Evaluation only \\
DiG-Bench \citep{DiG-bench}      & \gcmark & 1D string & Evaluation only \\
MazeBench \citep{Maze-bench}     & \rxmark & 3D render (also ASCII/JSON) & Evaluation only \\
gg-bench \citep{gg-bench}      & \rxmark & Text & Evaluation only (LLMs) \\
lmgame-Bench \citep{LMGame-bench}  & \rxmark & Frames or text & By game \\
Enigmata \citep{Enigmata}      & \rxmark & Text & By instance \\
HardcoreLogic \citep{HardCoreLogic}  & \rxmark & Text & Evaluation only \\
WITNESS (ours) & \gcmark & 2D ASCII board & By rule (validation): compositions, primitives \\
\bottomrule
\end{tabular}
}
\vspace{-1.5em}
\end{table}

\noindent\textbf{Learning and generalization through games.}
Beyond evaluation, games serve as RL environments, with the engine providing verifiable outcome rewards and levels serving as tasks.
Training on games has been reported to transfer beyond the training games in several settings.
\citet{Game-RL} and \citet{vision-zero} train vision-language models on game-derived tasks and report gains on multimodal benchmarks.
\citet{LMGame-bench} report that RL on Sokoban and Tetris improves performance on other games in the suite and on the planning tasks Blocksworld and WebShop.
\citet{Spiral} report gains of up to 10 points on eight reasoning benchmarks from self-play on zero-sum games, and \citet{gg-bench} train RL agents by self-play on generated games.
We study whether training language models on games with hidden rules improves performance on held-out compositions and held-out primitives, evaluating these two forms of transfer separately.

%We defer broader related work discussion on hypothesis testing, scientific discovery, and agent harnesses in Appendix~\ref{adxsec:deferred-related-work}.
%\vspace{-0.5em}

\noindent\textbf{Hypothesis testing and scientific discovery.}
The same loop of proposing a hypothesis, choosing an experiment, and revising the hypothesis from the outcome is studied under the framing of scientific discovery \citep{automation-of-science, where-science-starts}, and \citet{discovering-faster-mm} study how quickly models discover rules relative to their interaction budget.
We adopt four evaluations from this line as external transfer targets.
PhysGym \citep{PhysGym} evaluates the discovery of physical laws through interactive experiments, and WILT \citep{wilt} evaluates inductive rule inference from queried examples.
\citet{failing-to-falsify} evaluate whether models seek evidence that would falsify their own hypotheses, and CipherBank \citep{CipherBank} evaluates the inference of hidden encodings.
These evaluations probe related aspects of rule inference and hypothesis testing across domains and representations, allowing us to assess transfer beyond grid puzzles.

\noindent\textbf{Agent interfaces and harnesses.}
The interface between the model and the environment affects what an evaluation measures.
\citet{LMGame-bench} report that adding perception and memory modules raises the fraction of runs beating a random baseline from 60\% to 86.7\%.
ARC-AGI-3 results likewise depend on the agent scaffold around the model, and reported scores combine the model with scaffolds that differ across submissions \citep{arcagi3report}.
Frameworks that provide search or code execution, such as Prime-Agent \citep{karten2026prime} and [schema] \citep{schema2026}, introduce additional computational tools and interaction policies, so their scores are not directly comparable to those obtained under a fixed action-only interface.
A shared harness therefore controls one source of variation in cross-model comparisons.
Section~\ref{section:model-performance} evaluates all models under one harness, ablates its components to measure their effect, and reports results under such a framework as a reference point.
\section{Conclusion}
\vspace{-0.5em}
\label{section:conclusion}
We introduced WITNESS, a grid-based environment for studying interactive rule discovery, comprising WitnessBench for evaluation, WitnessGym for RL training, and WitnessForge for game generation.
With this environment we asked what limits current language models and whether RL improves performance.
On the first question, the highest-scoring model on RHAE-L5 under a shared harness solves only 24\% of private test level slots.
On the second question, RL training raises Qwen3.8-27B's private test RHAE-L5 from 2.1 to 5.4, improves performance on held-out compositions and held-out primitives, and yields an average gain of 4.1 percentage points across four external benchmarks, providing evidence of transfer beyond the training rules.
We hope this environment, composed of the challenging benchmark and generalizable RL training task, can foster further research in improving genuine reasoning intelligence of AI particularly in unfamiliar environments.

\newpage
\label{others}

\subsubsection*{Acknowledgments}
We thank all members of Fleet AI’s research and engineering teams for helpful discussions and constructive feedback.
We thank Neeraj Kashyap, Allie Gu, and Sam Weinberg for playing and recording the games, and Ventali Tan for porting WITNESS to the in-house platform.

%\newpage
\bibliography{references}
\bibliographystyle{preprint}
\newpage
\appendix
\section{APPENDIX: Ablations, Experiment Details, Qualitative Studies, Full Prompts, and Pipeline Pseudo-code}
\label{section:appendix}

\begin{figure}[h!]
    \centering
    \includegraphics[width=1\textwidth]{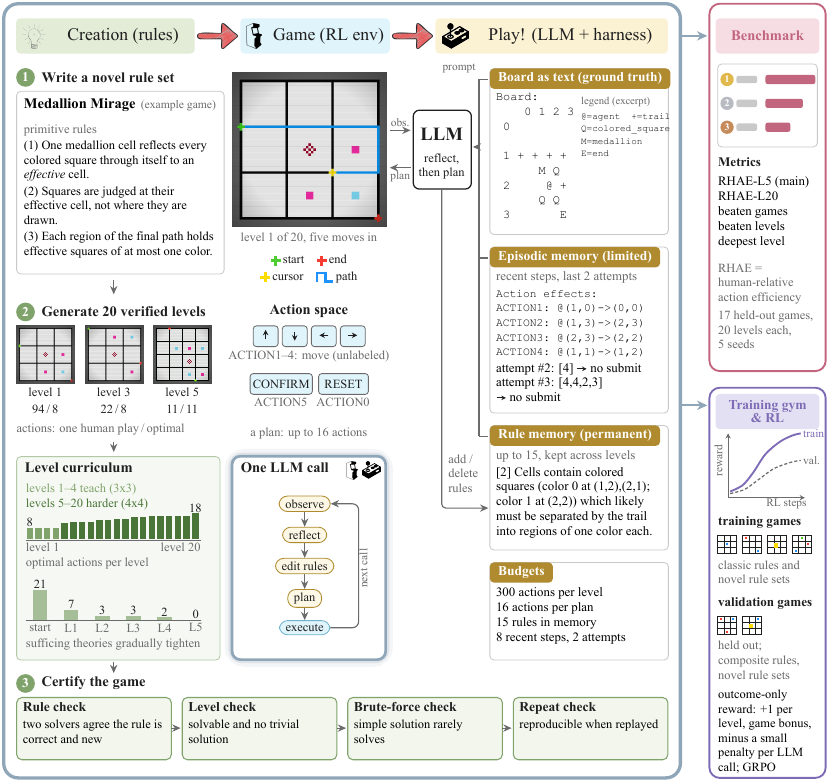}
    \caption{Cheatsheet for \textbf{W}itness-\textbf{I}nspired \textbf{T}estbed of \textbf{N}ovel \textbf{E}nvironments with \textbf{S}ymbolic \textbf{S}tructure (WITNESS).}
    \label{fig:fig1_cheatsheet}
\end{figure}

\subsection{Ablation: Given Ground-Truth Rules}
\label{section:gt-rule-ablation}

\begin{table}[t]\centering\footnotesize
\caption{Ground-truth (GT) rule cards as extra prompts for test-time ablation.
A ground-truth rule card is a hand-written natural-language statement of a game's true rules (win condition and mechanics, transcribed from the game engine), which we supply in the system prompt at evaluation time only, as an oracle probe that separates a policy's failure to discover the rules from its failure to act on them.
Eight held-out validation games, 5 evaluation seeds, 8192-token output cap, thinking off.
A rule card does not bring great improvements on the base model,
while the same card improves the RL policies significantly.
Frontier models such as Opus-5 are mainly bottle-necked by rule discovery.
These hint-assisted scores are only for diagnostic purpose and excluded from all other results.
}
\label{tab:gt-rule-card-probe}
\resizebox{\textwidth}{!}{
\begin{tabular}{llcccccc}\toprule
Policy & Extra prompt & RHAE-L5 & $\Delta$RHAE-L5 & RHAE-uncap & Games cleared & Efficiency & Plan-less calls \\
\midrule
Base (no RL) & none & 8.0 $\pm$ 5.0 & -- & 9.9 $\pm$ 6.7 & 5/40 & 0.39 & 1\% \\
 & generic text & 9.1 $\pm$ 2.5 & +1.1 & 11.4 $\pm$ 5.6 & 6/40 & 0.44 & 0\% \\
 & GT rule card & 10.1 $\pm$ 2.7 & +2.1 & 14.8 $\pm$ 7.4 & 6/40 & 0.48 & 1\% \\
\midrule
\begin{tabular}[c]{@{}l@{}}Broadcast RL, step 10\end{tabular} & none & 15.9 $\pm$ 3.4 & -- & 19.8 $\pm$ 2.1 & 7/40 & 0.50 & 2\% \\
 & generic text & 20.2 $\pm$ 4.8 & +4.3 & 22.3 $\pm$ 6.0 & 8/40 & 0.63 & 1\% \\
 & GT rule card & 47.7 $\pm$ 5.0 & +31.8 & 82.6 $\pm$ 14.8 & 31/40 & 0.69 & 57\% \\
\midrule
\begin{tabular}[c]{@{}l@{}}Credit-C RL, step 20\end{tabular} & none & 13.1 $\pm$ 3.5 & -- & 15.1 $\pm$ 3.5 & 7/40 & 0.50 & 12\% \\
 & generic text & 22.3 $\pm$ 7.8 & +9.2 & 25.8 $\pm$ 8.3 & 9/40 & 0.61 & 12\% \\
 & GT rule card & 48.4 $\pm$ 5.3 & +35.3 & 91.4 $\pm$ 25.7 & 29/40 & 0.71 & 71\% \\
\midrule
Opus 5 (frontier) & none & 59.9 $\pm$ 11.4 & -- & 152.8 $\pm$ 26.2 & 30/40 & 0.74 & 3\% \\
 & generic text & 63.6 $\pm$ 3.7 & +3.7 & 195.9 $\pm$ 25.5 & 34/40 & 0.72 & 4\% \\
 & GT rule card & 97.8 $\pm$ 0.7 & +37.8 & 356.2 $\pm$ 13.2 & 40/40 & 0.99 & 0\% \\
\midrule
GPT-6 Astra (frontier) & none & 63.7 $\pm$ 2.4 & -- & 194.2 $\pm$ 14.2 & 25/40 & 0.87 & 2\% \\
 & generic text & 68.8 $\pm$ 3.0 & +5.1 & 245.3 $\pm$ 22.6 & 30/40 & 0.81 & 1\% \\
 & GT rule card & 98.6 $\pm$ 0.0 & +34.9 & 366.0 $\pm$ 1.3 & 40/40 & 0.99 & 0\% \\
\bottomrule
\end{tabular}
}
\end{table}

To minimize the possibility of leakage of GT rules and contamination of the test set, we perform this ablation with the validation set.
For a frontier model such as Opus-5, with GT rules, it is able to solve all games and levels, and highly efficiently.
Given GT rules, RL-trained models tend to over-think with the rules and use up the context tokens before writing a plan. The rate of plan-less calls rises dramatically (e.g., from 2\% to 57\%), especially for more complex levels, although RL-trained models also achieve a much higher RHAE score (e.g., over +31\% improvement) with GT rule cards.

\begin{figure}[t]
    \centering
    \includegraphics[width=\textwidth]{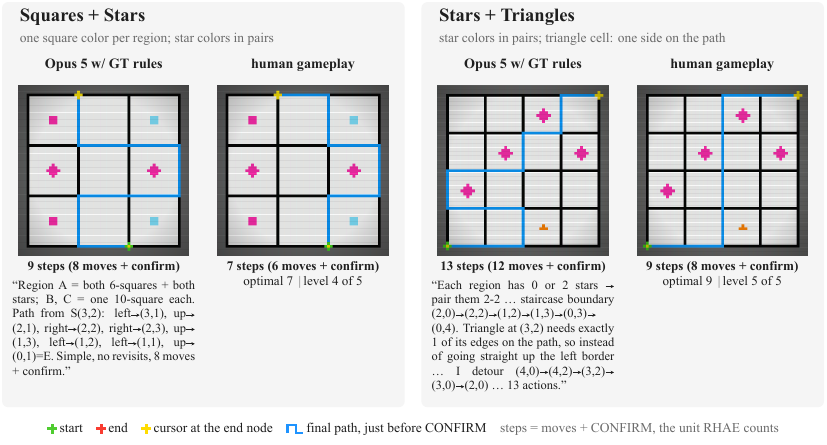}
    \caption{Why Opus-5 does not achieve perfect RHAE score (100) given GT rules: rare cases where human players complete a level more efficiently than Opus.
    In these levels, a human player selected a more efficient path, possibly unconsciously as it is a straight-forward path; the opus model selected non-optimal paths in 3/5 and 1/5 seeds, respectively.
    For most other cases, especially in earlier levels where a human player performed numerous steps to discover rules, Opus given GT rules executes much more efficiently, achieving RHAE-uncap scores that far exceed 100.}
    \label{fig:fig_gtcase_paths_row}
\end{figure}

\begin{promptbox}{Generic-text control:
additional text injected into the official eval harness as text content ablation for the GT rule card.
It answers the question of whether injecting additional guidance that does not reveal hidden rules is also helpful.
Length-matched to the GT rule cards.
}
The following is general guidance (given and correct). It states no rules of this game; the instruction elsewhere to rely only on observation still applies to the game itself.

\textbf{Action ids.} In this family of games ACTION1 = UP, ACTION2 = DOWN, ACTION3 = LEFT, ACTION4 = RIGHT, ACTION5 = CONFIRM (submits your work and moves nothing) and ACTION0 = RESET (restarts the current level; what you have learned is kept). Coordinates in the board text are written (row, col), 0-indexed, with (0,0) at the top-left.

\textbf{Observation.} Read the whole board text before acting: the legend, the Start and End lines, the Cell content list and the Path so far line describe the same board from different angles, so cross-check them against each other. Treat every symbol on the board as potentially meaningful until evidence shows otherwise, and note where symbols sit relative to each other and to the start and end. Prefer deliberate single-purpose tests over long speculative sequences; after each test compare the new board text with the previous one and record exactly what changed.

\textbf{Rules.} Keep rules short, testable and one sentence each; delete a rule as soon as an observation contradicts it instead of patching it with exceptions. Distinguish what you observed from what you inferred, and mark inferences as tentative until a second, independent observation supports them. If several rules could explain the same evidence, prefer the simplest one that explains every observation, and design the next action to tell the candidates apart.

\textbf{Efficiency.} Every action counts toward the step budget and the efficiency score, including moves that do nothing, failed confirms and resets; a level scores higher when it is solved in fewer actions, so think before acting and stop probing at random once the mechanics are clear.
\end{promptbox}

\subsection{Ablation: Level Diversity vs Rule Diversity}
\label{section:level-set-ablation}

We trained with games of fixed levels.
It is essential to observe whether we should add level diversity during training, i.e., training with the same games (identical rules) but different level sets (specific boards/tasks vary).
In Table \ref{tab:heldout-bank}, it shows that the gains in trained layouts carry over to the unseen layouts.
Therefore, level diversity is not necessary and we can focus on rule diversity.

\begin{table}[h!]\centering\small
\caption{Transfer of RL gains to unseen layouts of the same rules.
For each training game that has an alternate level set we evaluate checkpoints on the trained set and on a reserved set that was never used in training (same rules, different boards; the two layouts differ only in levels).
Cells: levels completed / action efficiency on completed levels / official game score.
The gains on trained layouts carry over to the unseen layouts;
the gap between trained and unseen does not grow with training.}
\label{tab:heldout-bank}
\resizebox{\textwidth}{!}{\begin{tabular}{llcccc}\toprule
Checkpoint & Layout & ft10 & ft14 & ft15 & ft16 \\
 & & ($\Delta$slots 4–5) & ($\Delta$slots 1–2) & ($\Delta$slots 0–4, 7) & ($\Delta$slots 4–7) \\ \midrule
Base (no RL) & trained & 1.19$\pm$0.32 / 0.29 / 0.6 & 2.25$\pm$0.37 / 0.46 / 3.0 & 2.81$\pm$0.32 / 0.45 / 3.4 & 2.06$\pm$0.46 / 0.59 / 4.5 \\
 & unseen & 0.75$\pm$0.37 / 0.38 / 1.1 & 1.31$\pm$0.38 / 0.41 / 1.9 & 2.12$\pm$0.38 / 0.43 / 2.7 & 3.69$\pm$0.38 / 0.40 / 4.5 \\[2pt]
RL, step 10 & trained & 6.44$\pm$0.80 / 0.67 / 40.0 & 3.12$\pm$0.40 / 0.50 / 5.8 & 2.50$\pm$0.48 / 0.59 / 4.6 & 4.31$\pm$0.30 / 0.48 / 8.2 \\
 & unseen & 5.94$\pm$0.63 / 0.64 / 25.2 & 2.88$\pm$0.48 / 0.41 / 4.9 & 2.62$\pm$0.38 / 0.44 / 3.9 & 4.62$\pm$0.34 / 0.62 / 13.7 \\[2pt]
RL, step 30 & trained & 3.75$\pm$0.81 / 0.52 / 12.9 & 3.44$\pm$0.27 / 0.51 / 7.0 & 1.88$\pm$0.32 / 0.58 / 2.7 & 2.44$\pm$0.46 / 0.58 / 5.3 \\
 & unseen & 5.00$\pm$0.80 / 0.53 / 21.1 & 2.12$\pm$0.41 / 0.31 / 2.8 & 2.31$\pm$0.38 / 0.38 / 2.4 & 3.25$\pm$0.50 / 0.59 / 8.5 \\[2pt]
\bottomrule
\end{tabular}}
\end{table}

\begin{figure}[t]
    \centering
    \includegraphics[width=1.0\textwidth]{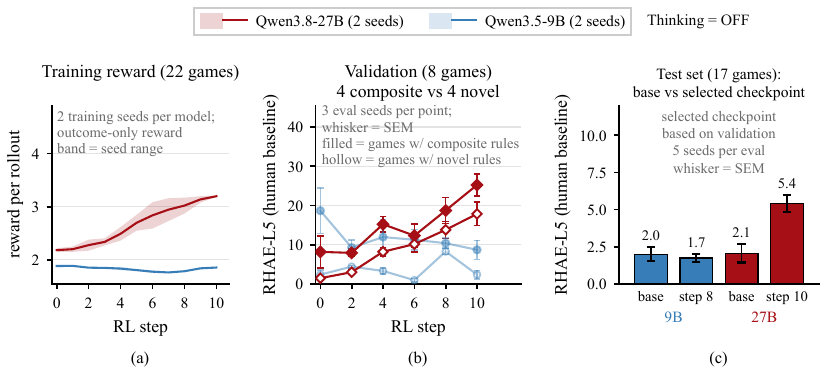}
    \caption{Comparing models of different sizes (9B vs 27B) with identical training settings.
    (a) 9B model does not experience useful learning signals while 27B learns to beat more levels. These findings corroborate the belief that complex behaviors emerge during model scaling;
    (b) the validation performance on composite of known primitive rules and unseen rules starts differently but converge. It suggests what model learns during RL training is generalized capability rather than memorizing specific patterns;
    (c) training on games of primitive rules generalized to test games of unseen rules.
    }
    \label{fig:fig3_scale_9b_vs_27b_rkae}
\end{figure}

\begin{figure}[t]
    \centering
    \includegraphics[width=1.0\textwidth]{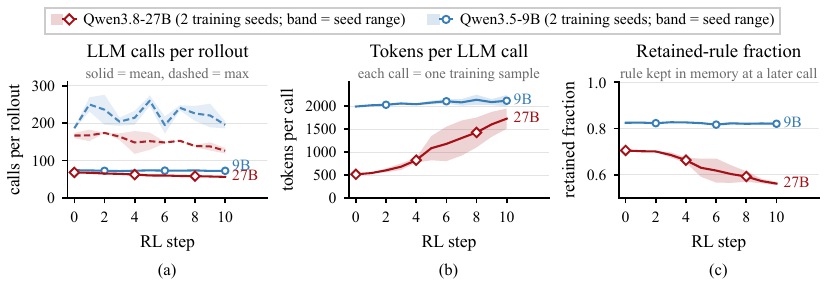}
    \caption{Behavior comparison between 9B and 27B models.
    (a) In contrast to 9B model, 27B model learns to complete more levels with less LLM calls within a game rollout as the training progresses;
    (b) 27B model learns to reflect more within each LLM call to plan high-quality actions, while 9B model consistently outputs verbose tokens that do not improve level-solving;
    (c) 27B model learns to preserve smaller fraction of rules it proposed earlier, while 9B model keeps generating and preserving most low-level rule hypotheses that do not contribute to puzzle solving.
    }
    \label{fig:fig5_behaviour_9b_vs_27b_survived}
\end{figure}

\subsection{Detailed RL Settings and Training Dynamics}
\label{section:rl-settings}

\textbf{Settings}.
We train Qwen3.5-9B and Qwen3.8-27B using Miles~\cite{miles2026} on B300 GPUs. At each training iteration, we sample 8 independent episodes for each of the 22 training games, giving 176 episodes before filtering. All episodes use the same set of levels during training and start from the first level. An episode ends when the agent completes all ten training levels (or all five, for the two training games that have only five levels), reaches 300 LLM calls, or exhausts the 300-action budget for an unsolved level.

At each call, the model receives the current ASCII board with its rule memory and recent attempts. It returns an analysis, memory operations, and an optional plan of up to 16 actions. The harness executes the plan before querying the model again. We sample at temperature 1 without specified top-$p$ or top-$k$ filtering and allow up to 8,192 response tokens per call. Native thinking mode is disabled. The model writes its analysis in the \texttt{<meta>} field required by the harness.

Under our 27B model training settings, it takes about 119 hours of wall-clock time on one $8\times B300$ node ($\approx 950$ GPU-hours) for the full 30-step training, with each step spending an average of 4.0 hours.
Because training is synchronous and the trainer and inference engine are co-located on the same GPUs, sampling and training proceed alternately;
the sampling stage, which waits for the slowest of the 176 rollouts, consumes about 75\% of total wall-clock.

\textbf{Reward}.
The agent receives a reward of 1 for each completed level and an additional 0.5 bonus for completing all ten levels (the two five-level games cannot earn this bonus). Each LLM call has a cost of 0.005. For an episode $\tau$, the return is
\begin{equation}
R(\tau)=n_{\mathrm{solved}}(\tau)
+0.5\,\mathds{1}[n_{\mathrm{solved}}(\tau)=10]
-0.005\,S(\tau),
\label{eq:witness-rl-return}
\end{equation}
where $S(\tau)$ counts all LLM calls in the episode, including calls whose responses are truncated or whose plans fail to advance the board. Level completion is checked by the game engine. We do not use a separate reward for the correctness of the model's explanations or proposed rules.

\paragraph{Rollout groups and advantage estimation.}
We group the sampled episodes by training game. Episodes interrupted by infrastructure failures are removed before computing advantages. Episodes that exhaust an action or LLM-call budget are retained. Let $\mathcal{B}$ denote the retained rollout batch and $\mathcal{G}_g$ its group of episodes for game $g$. Each sampled episode is counted separately, including episodes with identical trajectories.

For each nonempty group, we use its mean return as the baseline:
\begin{equation}
\small
\widetilde A(\tau)=R(\tau)-\frac{1}{|\mathcal{G}_g|}
\sum_{\tau'\in\mathcal{G}_g}R(\tau'),
\qquad \tau\in\mathcal{G}_g.
\label{eq:witness-group-centering}
\end{equation}
The advantages are centered and whitened once across the nonempty retained batch as following:
\begin{equation}
\small
\begin{split}
\mu_A&=\frac{1}{|\mathcal{B}|}
\sum_{\tau\in\mathcal{B}}\widetilde A(\tau),\\
\sigma_A^2&=\frac{1}{|\mathcal{B}|}
\sum_{\tau\in\mathcal{B}}\bigl(\widetilde A(\tau)-\mu_A\bigr)^2,\\
A(\tau)&=\begin{cases}
\displaystyle\frac{\widetilde A(\tau)-\mu_A}{\sigma_A+10^{-8}},
&\sigma_A\geq10^{-8},\\[4pt]
0,&\sigma_A<10^{-8}.
\end{cases}
\end{split}
\label{eq:witness-rl-advantage}
\end{equation}
Each episode contributes once to these statistics, regardless of its number of calls or tokens. We monitor both return statistics and the fraction of groups with varying solve counts.

\paragraph{Broadcast baseline.}
Let $A_j(\tau)$ be the advantage assigned to call $j$ in episode $\tau$. The broadcast baseline sets $A_j(\tau)=A(\tau)$ for every call and uses this value for all its retained response tokens.

\textbf{Policy objective}.
For call $j$, let $x_j$ and $y_j$ denote the stored prompt and response. Let $\pi_{\theta_{\mathrm{old}}}$ be the policy used to collect
the current batch. Writing $y_{j,t}$ for token $t$ of response $y_j$, the importance ratio is
\begin{equation}
\small
\rho_{j,t}(\theta)=
\frac{\pi_\theta(y_{j,t}\mid x_j,y_{j,<t})}
     {\pi_{\theta_{\mathrm{old}}}(y_{j,t}\mid x_j,y_{j,<t})}.
\label{eq:witness-rl-ratio}
\end{equation}
Here $y_{j,<t}$ denotes the response tokens preceding position $t$. We use the PPO clipped surrogate with lower and upper clipping thresholds of $1-\varepsilon_{\text{lo}}=0.8$ and $1+\varepsilon_{\text{hi}} = 1.28$. The token loss is
\begin{equation}
\small
\begin{split}
\ell_{j,t}(\theta;\tau)=-\min\!\Bigl\{
\rho_{j,t}(\theta)A_j(\tau), ~~~ \mathrm{clip}\bigl(\rho_{j,t}(\theta), 1-\varepsilon_{\text{lo}}, 1+\varepsilon_{\text{hi}}\bigr)A_j(\tau)
\Bigr\}.
\end{split}
\label{eq:witness-rl-token-loss}
\end{equation}
The old policy and advantages are held fixed during optimization. We use no KL penalty, entropy bonus, or dual clipping.

\paragraph{Token masking and loss aggregation.}
Responses that reach the output limit are truncated, but the episode continues. We exclude these responses from the policy loss.
Let $m_{j,t}=1$ for response tokens in retained calls and $m_{j,t}=0$ for tokens in truncated calls. Prompt and environment-observation tokens are not trained. Episodes contain different numbers of calls and response tokens. We first average the retained token losses within each call, then sum these per-call means over the batch, normalizing by the number of sampled episodes. For a non-empty batch in which every episode has at least one retained response token, we have
\begin{equation}
\small
\mathcal{L}(\theta)=\frac{1}{N}
\sum_{\tau\in\mathcal{B}}
\sum_{j}
\frac{\sum_{t}m_{j,t}\ell_{j,t}(\theta;\tau)}
     {\sum_{t}m_{j,t}}.
\label{eq:witness-rl-objective}
\end{equation}
Here $N$ is the trainer's fixed batch normalizer ($176$, the number of episodes sampled per iteration, in our configuration). The inner sum runs over the retained calls $j$ of episode $\tau$; both token sums run over the response tokens of call $j$; the episode dependence of $m_{j,t}$ and $\rho_{j,t}$ is implicit.
Only retained response tokens enter the numerator and denominator. Each retained call has equal weight in the batch average, so an episode's weight is proportional to its number of retained calls. Within a call, tokens share the call's weight equally.

For the broadcast baseline mentioned above, every call uses the same episode advantage, $A_j(\tau)=A(\tau)$. Substituting this into Equation~\ref{eq:witness-rl-objective} gives
\begin{equation}
\small
\begin{split}
\mathcal{L}_{\mathrm{broadcast}}(\theta)
={}&-\frac{1}{N}
\sum_{\tau\in\mathcal{B}}
\sum_{j}
\frac{1}{\sum_{t}m_{j,t}}
\sum_{t}m_{j,t} \cdot\min\!\Bigl\{
\rho_{j,t}(\theta)A(\tau),\,
\operatorname{clip}\bigl(
\rho_{j,t}(\theta),0.8,1.28
\bigr)A(\tau)
\Bigr\}.
\end{split}
\label{eq:witness-rl-broadcast-loss}
\end{equation}

\begin{figure}[t]
    \centering
    \includegraphics[width=0.8\textwidth]{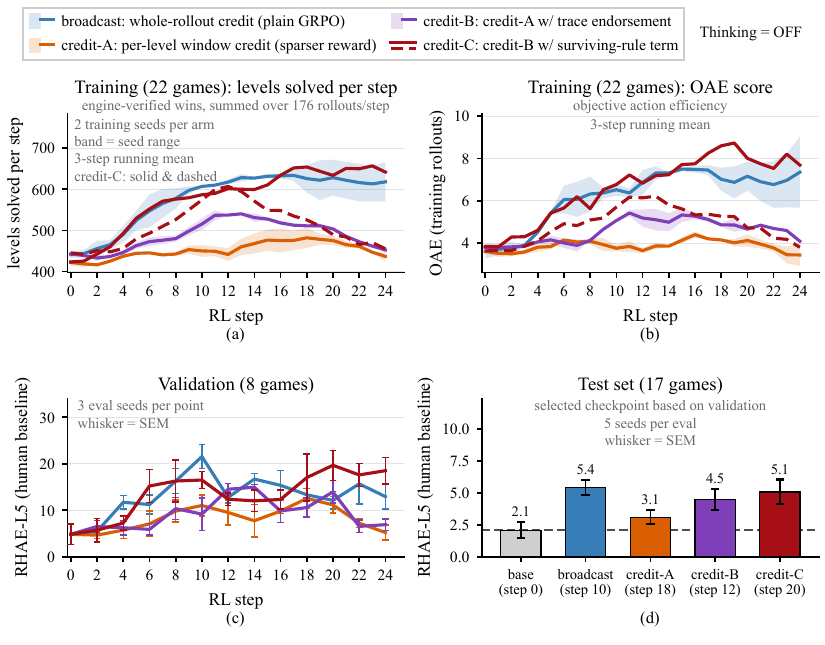}
    \caption{Comparison of RL algorithms: various credit assignment methods. Credit-C is shown with two curves, one with each seed, to exhibit its divergence during training, due to the noise in deterministically keyed rules.}
    \label{fig:fig4_rl_algo_ablation_rkae}
\end{figure}

\paragraph{Credit-assignment variants.}

The three variants compared in Section~\ref{section:training-gym} replace the broadcast advantage with the call-specific advantages. They use the same episode sampling, episode reward defined in Equation~\ref{eq:witness-rl-return}, clipped token loss in Equation~\ref{eq:witness-rl-token-loss}, and the call-mean loss reduction in Equation~\ref{eq:witness-rl-objective}. They differ in how the call advantages are computed and which calls enter the loss. An advantage of the call is shared by all its retained response tokens. For the credit variants, let $T_j=\sum_t m_{j,t}$ be the number of trainable response tokens in call $j$.

\paragraph{\textit{Credit-A: per-level-window credit.}}

For each episode $\tau$, we partition its calls into windows according to the level recorded for each call by the game engine. Let $W_{\tau,\ell}$ contain the calls assigned to level $\ell$, and let $W(j)$ be the window containing call $j$.

The local return used for window comparisons is
\begin{equation}
R(W)=\mathds{1}[\mathrm{solved}(W)] \bigl(1+0.5\,\mathds{1}[\mathrm{bonus}(W)]\bigr)
-0.005\,|W|,
\label{eq:witness-credit-window-return}
\end{equation}
where $|W|$ counts LLM calls, $\mathrm{solved}(W)$ indicates that the window solves its assigned level, and $\mathrm{bonus}(W)$ marks a window that completes the final level. This local return is distinct from
the episode return, whose cost is per environment action. We use $R(W)$ for the comparison below so a call that sweeps several levels contributes only one solved-window outcome to its sibling group.

For game $g$ and level $\ell$, let $\mathcal{W}_{g,\ell}$ contain the windows from the sampled episodes assigned to that game and level. We center each window return against its siblings:
\begin{equation}
\widetilde A(W)=R(W)-\frac{1}{|\mathcal{W}_{g,\ell}|} \sum_{W'\in\mathcal{W}_{g,\ell}}R(W'), ~~~ W\in\mathcal{W}_{g,\ell}.
\label{eq:witness-credit-sibling-centering}
\end{equation}
We remove all calls in a sibling group if it has fewer than two windows, every window is solved, or no window is solved. Such groups lack a contrast in level completion even if their call counts differ. Removed calls do not enter the policy loss or the normalization below. They are removed through the trainer's sample-removal hook rather than assigned zero advantage.

Let $\sigma$ be the standard deviation of $\widetilde A(W(j))$ across calls that remain trainable after this filter, counting the value once per call with $T_j>0$. When $\sigma>0$, Credit-A assigns
\begin{equation}
A_j^{\mathrm{A}}(\tau)=
\frac{\widetilde A\bigl(W(j)\bigr)}{\sigma}.
\label{eq:witness-credit-normalization}
\end{equation}
We divide by the batch-wide scale without subtracting the mean over calls. Sibling centering already sets the zero point for each group and a further mean over calls weights windows by their call counts.

Truncated calls remain in their windows when computing $R(W)$, which count in the window cost $0.005\,|W|$, and forming the sibling groups, including windows made entirely of truncated calls. They have $T_j=0$ and contribute neither to the loss nor to the calculation of $\sigma$.

\paragraph{\textit{Credit-B: crediting the calls that solved.}}

Credit-B uses the same windows, sibling centering, and scale $\sigma$ as Credit-A. It changes call weights only in solved windows with $\widetilde A(W)>0$, elsewhere $A_j^{\mathrm{B}}(\tau)= A_j^{\mathrm{A}}(\tau)$.

Within a level, an agent-chosen reset or a failed submit starts a new attempt. A failed submit is a confirm action that does not advance the level and clears the drawn path. The terminal give-up reset does not start an attempt. A call belongs to the attempt in which its plan began. If a boundary occurs during that plan, the new attempt starts with the next call.

Under Credit-B, a call in an eligible window is \emph{credited} if it belongs to the final attempt or chooses a reset. A failed submit only creates an attempt boundary and this event alone does not credit its call. For either variant $v\in\{\mathrm{B},\mathrm{C}\}$, a credited call receives raw weight 1 and every other call receives raw weight 0.2. Here \emph{credited} names the higher raw-weight class; the remaining calls still have non-zero raw weight. For a window with at least one trainable token, these weights are rescaled as
\begin{equation}
w_j^{(v)}=\widetilde w_j^{(v)}
\frac{\sum_{k\in W}T_k}{\sum_{k\in W}\widetilde w_k^{(v)}T_k}, \qquad
\sum_{j\in W}w_j^{(v)}T_j=\sum_{j\in W}T_j.
\label{eq:witness-credit-rescale}
\end{equation}
Here $\widetilde w_j^{(v)}$ denotes the raw weight assigned under variant $v$. In eligible windows, the call advantage is
\begin{equation}
A_j^{(v)}(\tau)=
w_j^{(v)}\frac{\widetilde A(W(j))}{\sigma}, \qquad v\in\{\mathrm{B},\mathrm{C}\}.
\label{eq:witness-credit-weighted-advantage}
\end{equation}
In other windows, both variants use $A_j^{\mathrm{A}}(\tau)$. The raw weights have a maximum relative ratio of five, and the rescaling preserves $\sum_{j\in W}w_j^{(v)}T_j$ in each window. Under the loss above, this is a token-mass constraint and it does not in general preserve the sum of call weights or the window's contribution to the loss. We compute $\sigma$ from the unweighted centered values, before applying the weights.

Below we discuss on Credit-C, which uses the same weighting rule and additionally endorses the surviving rules.

\paragraph{\textit{Credit-C: crediting rules that survive.}}

Credit-C also credits a call if it is the first to add a rule to memory and that rule appears in the union of verified rules from strictly later reflections in the episode. A rule can credit only its
first author. Although verification may occur later, the extra weight is applied to the authoring call in its original window.

The Credit-B and surviving-rule conditions combine by union: a call meeting several conditions still receives raw weight 1 only once. Credit-C uses the same eligibility condition, weight rescaling in
Equation~\ref{eq:witness-credit-rescale}, and unweighted scale $\sigma$ as Credit-B. The resulting call advantages from all three variants are used in Equation~\ref{eq:witness-rl-token-loss}.

\textbf{Optimization and schedule}.
We run 30 training iterations with one optimization epoch per rollout batch. We use Adam with a constant learning rate of $10^{-6}$, no warm-up, and a global gradient-norm limit of 1.0. The rollout engine receives the updated weights after each iteration.
We log importance-ratio statistics as a numerical health check: both log-probabilities in the ratio are computed by the training implementation; under on-policy updates the ratio deviates from 1 due to trainer-side numerical noise. We apply no importance-sampling correction for the residual mismatch between the sampling engine and the trainer.
Checkpoints are selected on the eight validation games. Validation uses 3 evaluation seeds, and the held-out test evaluation uses 5.

\subsection{OOD Generalization}
\label{section:appendix-ood-generalization}

\noindent\textbf{Transfer to external benchmarks.}
To test whether the gains extend beyond grid puzzles, we evaluate the base model and the checkpoints at iterations 10 and 20 on four external benchmarks of rule inference and hypothesis testing (Section~\ref{section:related-work}), all with thinking on and a 32k-token cap.
Table~\ref{tab:dose_response_2x4} reports each score with its paired difference to the base model and a paired-bootstrap 95\% confidence interval.
At iteration 20, the broadcast checkpoint improves the mean over the four benchmarks by 4.1 points (95\% CI 2.1 to 6.2), with per-benchmark intervals excluding zero for CipherBank (+4.8) and WILT (+2.3), a lower bound at zero for PhysGym (+3.4), and an interval including zero for Failing-to-Falsify (+6.1).
The credit-C checkpoint at iteration 20 improves the mean by 2.7 points (95\% CI 0.6 to 4.9).
At iteration 10, the mean improvements of both checkpoints are smaller and their intervals include zero.
These results indicate that part of what is learned on WitnessGym transfers to rule inference outside the suite, with the clearest gains on cipher decryption and experiment design; the size of the effect is modest and the per-benchmark evidence is uneven.

\providecommand{\dcell}[2]{\begin{tabular}[c]{@{}c@{}}#1\\#2\end{tabular}}
\begin{table}[t]
\centering\small\caption{Training dynamics of OOD generalization:
credit assignment (broadcast vs.\ credit-C) $\times$ training step (10 vs.\ 20) on the four benchmarks.
During RL training, both models have progressive OOD gains against the base model.
}
\label{tab:dose_response_2x4}
\resizebox{\linewidth}{!}{\providecommand{\dcell}[2]{\begin{tabular}[c]{@{}c@{}}#1\\#2\end{tabular}}
\providecommand{\dcell}[2]{\begin{tabular}[c]{@{}c@{}}#1\\#2\end{tabular}}
\definecolor{deltagreen}{RGB}{34,139,34}
\providecommand{\dg}[1]{\textcolor{deltagreen}{(#1)}}
\providecommand{\dgm}[1]{\textcolor{deltagreen}{#1}}
\begin{tabular}{l c c c c c c}
\toprule
\textbf{Model} & \textbf{Step} &
\begin{tabular}[c]{@{}c@{}}\textbf{CipherBank}\\\textbf{custom ciphers (\%)}\\cipher decryption\\786 items, 2 reps\end{tabular} &
\begin{tabular}[c]{@{}c@{}}\textbf{PhysGym L4}\\\textbf{success (\%)}\\law discovery\\97 problems, 3 reps\end{tabular} &
\begin{tabular}[c]{@{}c@{}}\textbf{WILT}\\\textbf{elim@5 (\%)}\\rule discovery\\50 rules, 5 reps\end{tabular} &
\begin{tabular}[c]{@{}c@{}}\textbf{Failing-to-Falsify}\\\textbf{incompat.\ fraction (\%)}\\Wason 2-4-6\\80 instances, 5 reps\end{tabular} &
\begin{tabular}[c]{@{}c@{}}\textbf{Mean $\Delta$}\\\textbf{over 4 benchmarks}\\{[95\% CI]}\end{tabular} \\
\midrule
Qwen3.8-27B (base) & -- & 73.8 $\pm$ 0.2 & 44.7 $\pm$ 1.6 & 86.3 $\pm$ 1.6 & \dcell{50.7 $\pm$ 5.1}{\footnotesize paired subset: 49.3--49.6} & -- \\
\midrule
\quad + RL broadcast & 10 & \dcell{77.6 $\pm$ 2.3}{\dg{+3.8}} & \dcell{45.4 $\pm$ 1.8}{\dg{+0.7}} & \dcell{87.8 $\pm$ 1.1}{\dg{+1.5}} & \dcell{51.2 $\pm$ 3.5}{\dg{+0.5}} & \dgm{+1.6} [$-$0.3, +3.6] \\[2pt]
                     & 20 & \dcell{78.6 $\pm$ 1.3}{\dg{+4.8}} & \dcell{48.1 $\pm$ 2.1}{\dg{+3.4}} & \dcell{88.6 $\pm$ 1.7}{\dg{+2.3}} & \dcell{55.4 $\pm$ 4.9}{\dg{+6.1}} & \dgm{\textbf{+4.1}} [+2.1, +6.2] \\
\midrule
\quad + RL credit-C  & 10 & \dcell{77.0 $\pm$ 1.4}{\dg{+3.2}} & \dcell{45.4 $\pm$ 2.7}{\dg{+0.7}} & \dcell{87.6 $\pm$ 1.8}{\dg{+1.3}} & \dcell{51.8 $\pm$ 3.6}{\dg{+1.2}} & \dgm{+1.6} [$-$0.6, +3.8] \\[2pt]
                     & 20 & \dcell{78.1 $\pm$ 0.7}{\dg{+4.3}} & \dcell{46.4 $\pm$ 1.0}{\dg{+1.7}} & \dcell{87.5 $\pm$ 1.0}{\dg{+1.2}} & \dcell{54.0 $\pm$ 3.1}{\dg{+3.7}} & \dgm{\textbf{+2.7}} [+0.6, +4.9] \\
\bottomrule
\end{tabular}
}
\end{table}

\subsection{Mis-behaviors in OSS Harness}
\label{section:oss-harness-mechanism}

\textbf{Harness and evaluation set}.
The canonical evaluation harness presents the model with a human-readable board state. The model submits one action per turn, and each \texttt{CONFIRM} or \texttt{RESET} counts as a turn. Code execution and programmatic search are strictly prohibited. In the \emph{OSS harness: Prime-Agent} configuration (Table~\ref{table:harness_ablation}), the agent instead receives raw $64\times64$ image frames and has access to a persistent Python REPL and an HTTP interface to the environment referee. These interfaces allow code to parse frames, submit actions, and use referee feedback to test candidate paths. We rescore the Prime-Agent rollouts to distinguish level completion from the direct rule inference and action selection evaluated by the canonical harness.

The evaluation covers 17 private test games. It includes 34 rollouts each for Opus-5 and 4.8, and 85 rollouts each for Kimi-K3 and Qwen3.8-27B. Here \emph{mis-behavior} denotes a departure from the benchmark's intended measurement construct, not necessarily meaning a violation of the coding agent's system instructions.

\paragraph{Model-attended scoring.}
The model-attended score requires model participation in each credited level. We exclude a level completion if it meets any of the following criteria:
\begin{itemize}
    \item \textbf{Unattended Play (N):} No model forward pass occurs between the previous level completion and the current one. For example, a script solves the next level without the model observing its state or choosing its moves.
    \item \textbf{Programmatic Brute-Forcing (B1):} One code-cell execution produces three or more failed \texttt{CONFIRM} submissions. This threshold was set before rescoring. One or two failures in a cell do not trigger B1.
    \item \textbf{Background Execution (B2):} A background thread continues to submit actions after the code cell that started it has returned.
\end{itemize}
\textbf{N} concerns whether the model was called again before a level was completed. \textbf{B1} concerns repeated failed submissions within a single code-cell execution; failures from separate cells are not combined to reach its threshold. \textbf{B2} concerns actions submitted after control has returned from the initiating cell. For example, if a model starts a loop that clears several consecutive levels, the first completion can receive credit only if it satisfies the other criteria. Later completions without another model forward pass are excluded under N. A cell that repeatedly submits paths and resets after failures until one succeeds meets B1 once it produces three failed \texttt{CONFIRM} submissions. These cases differ from the model observing a failure and selecting the next attempt itself.

\paragraph{Verbal-only scoring.}
The verbal-only score applies the model-attended exclusions and adds one more criterion. We exclude a level under \textbf{S} if code generates its winning action sequence through search, enumeration, empirical rule-fitting, or reuse of a solver. Copying a path printed by code also triggers \textbf{S}. The exclusion applies even if the model wrote the solver or proposed the underlying rule: the program selected the moves used to complete the level.

We trace the provenance of each submitted action list. A level receives verbal-only credit if and only if the model explicitly typed the winning sequence, did not take it from prior code output or a programmatic solver, and the typed sequence matches the submitted moves. Code used solely to parse image frames is permitted under both scores, since the canonical harness already supplies parsed, human-readable boards.

\paragraph{Cascading Penalties (Rule B).}

For each score, we identify the first excluded level in a rollout. Under the preregistered strict penalty, Rule B, that level and all subsequent levels receive zero credit. A later level may therefore receive zero even if its own actions were model-attended. The rule prevents later completions from receiving credit when they may have benefited from rules or solvers acquired on an excluded level.

\paragraph{Environment probing and rollout termination.}
\label{app:prime-behaviours:cases}
Prime-Agent models used their tools to parse image frames, retain solvers in memory, and run scripts across levels. We also observed models inspecting local directories, searching system files for game rules or source code, listing active processes and ports, querying undocumented referee endpoints, and attempting network requests or package installations. None of these probes successfully retrieved privileged game information. An attempted probe alone therefore does not exclude a level; exclusions follow the criteria above.

Rollouts also encountered failed context summaries, timeout errors, dead kernels, and token or session limits. The reported scores reflect the point at which each rollout terminated.

\subsection{Prompts}
\label{section:prompts}

\begin{promptbox}{System prompt (identical for every LLM call of a game, up to a per-call ``Known Action Semantics'' block summarizing the agent's own observations that the harness appends; the action placeholders are resolved to the game's action set, ACTION1--5 for all Witness games, and the plan cap to 16). Besides the coordinate convention in P4, the three trailing paragraphs are the only game-mechanic facts the agent is told: which button confirms, what RESET does, and the plan-length cap.}
You are an interactive reasoning agent playing a puzzle video game. Each turn you receive a TEXT observation: an ASCII board with a legend and, on games that expose them, ground-truth lists of the board's markers and cell content, plus your recent actions with their effects and your own knowledge rules. Your goal: discover the rules and solve each level efficiently.

\textbf{\#\# Core Priors}

P1 OBJECTNESS: The environment has discrete bounded objects (the board elements named in the legend and, when present, the cell-content list). Objects persist between steps unless explicitly changed. They have: color, position, size, shape. Objects can appear, disappear, or change --- track these transitions.

P2 AGENTNESS: You control something in the environment. Your actions (ACTION1-ACTION5) cause state changes. You must discover WHAT each action does by observing before/after. Goal: reach a winning state (the environment will signal level completion).

P3 NUMBERS: Objects can be counted. Quantities may be constraints (exactly N, at least N, at most N). Arithmetic relationships between quantities may matter.

P4 GEOMETRY: Spatial relationships matter --- adjacency, containment, alignment, symmetry (horizontal/vertical/rotational). Distances can be Manhattan or Euclidean. Boundaries may block, wrap, or reflect. Coordinate system: (row, col), 0-indexed, (0,0) at TOP-LEFT --- the same order as every coordinate in the observation. On a dot board the cursor moves on dot intersections; cell (r,c) is the square between dots (r,c), (r,c+1), (r+1,c) and (r+1,c+1).

\textbf{\#\# Hypothesis-Test Protocol}

Each turn:\\

1. OBSERVE: Read the current state carefully\\
2. HYPOTHESIZE: What rule explains the observations so far?\\
3. PREDICT: What will my chosen action cause?\\
4. ACT: Choose the most informative action\\
5. VERIFY: Was prediction correct? Update hypothesis confidence.

\textbf{\#\# Exploration Strategy}

- Simple first: try basic actions before complex strategies\\
- Systematic: test each available action from new states\\
- Contrastive: different actions in the same state reveal what each action does\\
- Errors are signals: negative feedback (visual changes, state resets) = direct evidence of constraint boundaries\\
- Hypothesis-driven: once you have a hypothesis, test it efficiently\\
- AVOID CYCLES: if you've been to a state before, try a DIFFERENT action\\
- Map actions early: in the first few steps, try each of ACTION1-4 to learn what they do

\textbf{\#\# Reasoning Principles}

- Occam's Razor: prefer the simplest hypothesis that explains all observations\\
- One rule at a time: test hypotheses individually, not in combination\\
- Distinguish mechanics from goals: HOW things work vs WHAT you need to achieve\\
- Failed attempts are the most informative: they reveal constraint boundaries\\
- Look for patterns across levels: rules often stay the same, only configurations change

\textbf{\#\# HARD RULES --- Mandatory Constraints}

These rules are ABSOLUTE. Violating any of them wastes budget and degrades performance.

\textbf{\#\#\# Rule Discovery}\\

- R1: Output AT MOST 2 new rules per reflection. Quality over quantity.\\
- R2: DELETE any rule where counter-evidence $\geq$ supporting evidence. Do not keep disputed rules alive.\\
- R3: Rules below 0.2 confidence are dead --- stop referencing them.\\
- R4: Maintain AT MOST 15 active rules. If at capacity, delete the weakest before adding.\\
- R5: NEVER create a rule from a single observation. Require at least 2 supporting observations.\\
- R6: Correlation is NOT causation. ``X happened after Y'' does not mean Y caused X --- verify with a deliberate test.\\
- R7: If uncertain about a rule, cap its confidence at 0.5.

\textbf{\#\#\# Planning}\\

- P1: Plans must be AT MOST 16 actions. Longer plans accumulate prediction errors.\\
- P2: Each action in a plan MUST have a justification grounded in a known rule.\\
- P3: If 3 consecutive actions in a plan result in noop, ABANDON the plan immediately.

\textbf{\#\#\# Global}\\

- G1: When budget $>$ 95\% consumed, STOP exploring and crystallize knowledge immediately.\\
- G2: If 10 consecutive actions produce no state change, SWITCH strategy (different actions, different region).\\
- G3: When the same (state, action) produces different results, this is COGNITIVE DISSONANCE --- trigger immediate reflection to resolve the contradiction.

ACTION5 is the CONFIRM/SUBMIT action: it does not move anything --- press it to submit your COMPLETED solution and win the level. Plan it as the FINAL action, and only once your path satisfies every rule (confirming an incomplete/incorrect path fails the attempt).

ACTION0 is RESET: it restarts the CURRENT level from scratch. Your drawn path/progress on the board is erased, but everything you have LEARNED (rules, action semantics) is kept. IMPORTANT: all actions you have already taken on this level still count toward your step budget and efficiency score --- resetting does not refund them. Use it deliberately, when your current path is unsalvageable and restarting is cheaper than undoing it move by move.

PLAN LENGTH: at most 16 actions per plan are executed; anything beyond the first 16 is dropped. Plan in chunks of at most 16 actions and continue from the resulting state.
\end{promptbox}

\begin{promptbox}{User prompt of every LLM call (one call = one reflection: analyze, edit the rule memory, optionally emit a plan of up to 16 actions that the harness then executes). Slots in angle brackets are filled by the harness from the engine's ground truth and the agent's own memory; nothing in them reveals a hidden rule. The `Harness notices'' block appears only when the harness ignored or altered one of the agent's operations;` Recent Attempts'' once a plan has finished on the current level (harness trial memory); the warning block only when a rule has conflicting evidence.}
You are analyzing an unknown game environment to discover its rules and mechanics.\\
You must figure out everything from observation --- no prior knowledge of this game exists.

\textbf{\#\# Current Board State}\\
$\langle$compact board text: grid type and size, `Agent at row=$r$, col=$c$'',` Level: $k$/$M$'', the coordinate convention, a legend naming each symbol on the board, the ASCII board, Start/End coordinates, the mandatory dots and whether each is visited, on games that expose them the cell contents (e.g.\ square colours), `Path so far'', the per-action test status (`ACTION1: untested'', `ACTION4 (RIGHT): noop=0\% [observed $\times$3 this level]'' or` ACTION5 (CONFIRM/SUBMIT): pressed $\times$1 this level, 1 rejected''), and ``Last change: agent moved from $(r_1,c_1)$ to $(r_2,c_2)$'' (or the outcome of a CONFIRM/RESET)$\rangle$

\textbf{\#\# Known Action Semantics}\\
$\langle$per action, the direction, no-op rate and observation count the agent has itself established, plus the CONFIRM press count$\rangle$

\textbf{\#\# Recent Observations (last $n$ steps)}\\
$\langle$`No informative observations yet.'' or: the most recent observed movement per action, followed by the last actions (up to 8) and their effects, e.g.\` \#3 ACTION5: submit REJECTED --- path reset to start (2,0)''; a step the harness drew at random after a planless reflection is labelled as such$\rangle$

\textbf{\#\# Current Knowledge ($n$/15 rules)}\\
$\langle$`No rules discovered yet.'' or the agent's numbered rules` [i] $\langle$rule text$\rangle$'', a rule with confidence $\geq 0.8$ carrying a \(\checkmark\), a rule with conflicting evidence carrying {\fontencoding{U}\fontfamily{futs}\selectfont\char 49\relax}DISPUTED, a rule kept from the previous level carrying [carried]$\rangle$

\textit{[only when the harness ignored or altered one of the agent's operations]}\\
\textbf{\#\# Harness notices (since your last reflection)}\\
$\langle$one line per event, e.g.\ ``plan cap: your plan had 18 actions; only the first 16 ran --- the last 2 were not executed''$\rangle$

\textit{[once a plan has finished on this level]}\\
\textbf{\#\# Recent Attempts (this level)}\\
You have made $\langle n\rangle$ attempts on this level so far.\\

- attempt \#$\langle i\rangle$: [$\langle$action ids$\rangle$] $\to$ $\langle$end reason, e.g.\ `submit rejected at action 3/3'' or` no submit --- wandered without attempting; included 1 reset(s)''$\rangle$ \quad (last two attempts; a plan that ended without CONFIRM or RESET is listed as a ``chunk'' and not counted)\\
\textit{[if the last two attempts were the same actions from the same state]} {\fontencoding{U}\fontfamily{futs}\selectfont\char 49\relax} Your last two attempts were IDENTICAL action sequences from the same starting state.\\
(Attempt history is provided automatically --- do not store attempts as knowledge rules.)

\textbf{\#\# Instructions}\\
Analyze the recent observations in context of the board state and existing knowledge.\\
Look for:\\

- How actions affect the game state (movement patterns, special behaviors)\\
- Spatial patterns (board structure, special positions, boundaries)\\
- Win condition clues (what might lead to level completion)\\
- Effective strategies (sequences that make progress toward goals)

\textit{[only when a rule is disputed]} {\fontencoding{U}\fontfamily{futs}\selectfont\char 49\relax} ATTENTION --- The following rules have CONFLICTING evidence and may be wrong. Review carefully and consider deleting:\\
\hspace*{1em}Rule [$\langle i\rangle$]: ``$\langle$rule text$\rangle$'' (support=$\langle a\rangle$, contradictions=$\langle b\rangle$)

Respond with:\\

1. \texttt{\textless meta\textgreater}Your concise analysis\texttt{\textless/meta\textgreater}\\
2. Memory operations (one or more of these):\\
\hspace*{1em}\texttt{\textless add\textgreater}new rule or pattern you discovered\texttt{\textless/add\textgreater}\\
\hspace*{1em}\texttt{\textless delete\textgreater}rule\_number\texttt{\textless/delete\textgreater} \quad (to remove an outdated/wrong rule)\\
\hspace*{1em}\texttt{\textless keep/\textgreater} \quad (if no changes to knowledge needed)\\
3. Optionally, if you have a specific strategy to try right now:\\
\hspace*{1em}Use \texttt{\textless plan\textgreater}...\texttt{\textless/plan\textgreater} with action IDs as comma-separated numbers.

Reminders (full constraints in system prompt --- R1-R7, P1-P3, G1-G3):\\

- Update knowledge via delete+add (don't duplicate existing rules); keep rules one sentence each.\\
- DISPUTED rules ({\fontencoding{U}\fontfamily{futs}\selectfont\char 49\relax}) have conflicting evidence --- verify before relying.
\end{promptbox}

\subsection{Evaluation Details}
\label{section:appendix-eval}

Detailed evaluation settings including exact model version identifiers and reasoning budgets are given in Table~\ref{tab:model_provenance}

\begin{table}[t]
\centering\scriptsize\setlength{\tabcolsep}{3pt}\renewcommand{\arraystretch}{1.08}
\caption{\textbf{Exact model identifiers and sampling settings behind Table~\ref{tab:model_evals}.}
Every model was run with the same canonical harness and protocol:
17 unseen games $\times$ 5 evaluation seeds $\times$ up to 20 levels;
300 actions per level (an unsolved level ends the rollout) and 1{,}000 LLM calls per rollout; \texttt{temperature}\,$=1.0$ for all models (the client forces 1.0 whenever reasoning is enabled), no \texttt{top\_p}, \texttt{top\_k} or seed parameter, cached system prompt.
\emph{Reasoning}: ``max'' sends \texttt{reasoning:\{effort:"max"\}};
``16k budget'' sends \texttt{reasoning:\{max\_tokens:16000\}} (endpoints without effort tiers); ``adaptive-max'' is the Anthropic API's adaptive thinking with \texttt{output\_config.effort\,=\,max}.
\emph{Out cap} is the per-call output-token limit.
}
\label{tab:model_provenance}
\begin{tabular}{@{}l >{\raggedright\arraybackslash}p{130pt} l l@{}}
\toprule
Model & Requested identifier & Out cap / reasoning & Evaluated (2026) \\
\midrule
Fable-5            & \texttt{claude-fable-5}                          & 64k / adaptive-max & Jul 31 -- Aug 5 \\
Opus-5             & \texttt{anthropic/\allowbreak claude-opus-5}     & 32k / 16k budget   & Jul 29 -- Aug 5 \\
GPT-6-Astra-Pro    & \texttt{openai/\allowbreak gpt-6-astra-pro}      & 64k / max          & Sep 9 -- Sep 17 \\
GPT-6-Astra        & \texttt{openai/\allowbreak gpt-6-astra}          & 64k / max          & Sep 4 -- Sep 12 \\
Kimi-K3            & \texttt{moonshotai/\allowbreak kimi-k3}          & 32k / max          & Jul 30 -- Aug 6 \\
GPT-5.6-Sol        & \texttt{openai/\allowbreak gpt-5.6-sol}          & 32k / max          & Aug 12 -- Aug 22 \\
Opus-4.8           & \texttt{anthropic/\allowbreak claude-opus-4.8}   & 64k / max          & Jul 30 -- Aug 5 \\
Muse-Spark-1.2     & \texttt{meta/\allowbreak muse-spark-1.2}         & 32k / max          & Aug 11 -- Aug 19 \\
Grok-4.6           & \texttt{x-ai/\allowbreak grok-4.6}               & 32k / max          & Aug 12 -- Aug 17 \\
DeepSeek V4 Pro    & \texttt{deepseek/\allowbreak deepseek-v4-pro-0813} & 32k / max        & Aug 14 -- Aug 19 \\
Qwen-3.8-Max       & \texttt{qwen/\allowbreak qwen3.8-2.4t-a95b}      & 32k / max          & Aug 14 -- Sep 4 \\
Qwen3.8-27B        & \texttt{qwen/\allowbreak qwen3.8-27b}            & 32k / max          & Aug 17 -- Aug 20 \\
GLM-5.2-Max        & \texttt{z-ai/\allowbreak glm-5.2}                & 32k / max          & Jul 29 -- Aug 5 \\
GLM-5.3            & \texttt{z-ai/\allowbreak glm-5.3}                & 32k / max          & Aug 21 -- Aug 25 \\
Qwen3.5-35B-A3B    & \texttt{qwen/\allowbreak qwen3.5-35b-a3b}        & 32k / 16k budget   & Jul 29 -- Aug 4 \\
Glim-30            & \texttt{meta/\allowbreak muse-glimmer-30b}       & 32k / max          & Aug 11 -- Aug 12 \\
Dots-3-Note-Preview& \texttt{dots-studio/\allowbreak dots-3-note-preview:free} & 32k / 16k budget & Aug 16 -- Aug 17 \\
Qwen3.5-9B         & \texttt{qwen/\allowbreak qwen3.5-9b}             & 32k / 16k budget   & Jul 29 -- Aug 4 \\
\bottomrule
\end{tabular}
\vspace{2pt}
\end{table}

\subsection{Pipeline}
\label{section:pipeline}

The full game creation pipeline is given in Algorithm~\ref{alg:game_pipeline}.

\begin{algorithm}[!h]
\caption{Game Creation Pipeline}
\label{alg:game_pipeline}
\begin{algorithmic}
\small
\Require Natural-language rule concept
\Ensure Certified and packaged game bundle

\vspace{0.4em}
\State \textbf{Phase 1: Specification \& Implementation}
\State Generate natural-language spec with edge cases and pass neutrality tests.
\State $I_1, I_2 \leftarrow$ Independently generate two implementations from the spec.
\State Cross-examine $I_1$ and $I_2$ across a suite of test boards.
\If{$I_1 \neq I_2$}
    \State Generate counter-examples and retry (up to 4 rounds) until zero disagreements.
\EndIf
\State Merge $I_1$ and $I_2$ into a single certified module $M$.

\vspace{0.4em}
\State \textbf{Phase 2: Qualification Exams}
\State Verify $M$ for: necessity, visibility, non-triviality, and novelty against existing rules.

\vspace{0.4em}
\State \textbf{Phase 3: Level Generation \& Filtering}
\State $L_{\text{cand}} \leftarrow$ Generate a large pool of candidate levels.
\State $L_{\text{valid}} \leftarrow \emptyset$
\For{each level $l \in L_{\text{cand}}$}
    \State \textbf{Require:} $l$ passes structural validation and optimal solution depth $\ge 5$
    \State \textbf{Require:} $l$ necessitates rule mechanics (no heuristic shortcuts)
    \State \textbf{Require:} Win rate is $0$ for all greedy baselines and $\le 0.004$ for random play
    \If{$l$ satisfies all requirements}
        \State $L_{\text{valid}} \leftarrow L_{\text{valid}} \cup \{l\}$ \Comment{De-duplicated via canonical hash}
    \EndIf
\EndFor

\vspace{0.4em}
\State \textbf{Phase 4: Sequence Assembly \& Teaching}
\State Measure and profile kill-able wrong theories across $L_{\text{valid}}$.
\State Schedule levels $S$ monotonically: Introduce $\rightarrow$ Accumulate $\rightarrow$ Reinforce.
\State Verify $S$ against the shipped bank for schedule conformance and feature monotonicity.
\State Embed visual feedback and exploration constraints.

\vspace{0.4em}
\State \textbf{Phase 5: Audit \& Packaging}
\State Conduct human blind-judging and adversarial algorithmic testing.
\State Bundle $M$ and $S$ into the final package.
\State Execute canonical sequential replay to confirm deterministic, hash-checked victories.
\State \Return Game Bundle
\end{algorithmic}
\end{algorithm}

\end{document}